\documentclass[algorithms,article,accept,oneauthor]{style/mdpi}

\usepackage{xfrac}

\firstpage{1}
\pubvolume{1}
\issuenum{1}
\articlenumber{0}
\pubyear{2023}
\copyrightyear{2023}
\datereceived{ }
\daterevised{ } 
\dateaccepted{ }
\datepublished{ }
\hreflink{https://doi.org/} 

\Title{Entropy and the Kullback-Leibler Divergence for Bayesian Networks:
  Computational Complexity and Efficient Implementation}
\TitleCitation{Entropy and the Kullback-Leibler Divergence for Bayesian Networks}

\Author{Marco Scutari $^{1}$}
\AuthorNames{Marco Scutari}
\AuthorCitation{Scutari, M.}
\address{$^{1}$ \quad Istituto Dalle Molle di Studi sull'Intelligenza
  Artificiale (IDSIA), Lugano, Switzerland; scutari@bnlearn.com}
\corres{Correspondence: scutari@bnlearn.com}

\abstract{
  Bayesian networks (BNs) are a foundational model in machine learning and
  causal inference. Their graphical structure can handle high-dimensional
  problems, divide them into a sparse collection of smaller ones, underlies
  Judea Pearl's causality, and determines their explainability and
  interpretability. Despite their popularity, there are almost no resources in
  the literature on how to compute Shannon's entropy and the Kullback-Leibler
  (KL) divergence for BNs under their most common distributional assumptions. In
  this paper, we provide computationally efficient algorithms for both by
  leveraging BNs' graphical structure, and we illustrate them with a complete
  set of numerical examples. In the process, we show it is possible to reduce
  the computational complexity of KL from cubic to quadratic for Gaussian BNs.
}


\DeclareMathOperator{\tr}{tr}

\DeclareMathOperator{\rarr}{\rightarrow}
\DeclareMathOperator{\uarr}{{\relbar\mkern-9mu\relbar}}

\begin{document}

\newcommand{\mref}[1]{(\ref{#1})}
\newcommand{\Prob}[1]{\operatorname{P}\left(#1\right)}
\newcommand{\E}{\operatorname{E}}
\newcommand{\VAR}{\operatorname{VAR}}
\newcommand{\COV}{\operatorname{COV}}
\newcommand{\COR}{\operatorname{COR}}
\newcommand{\given}{\mid}
\newcommand{\indep}{\hspace{0.1em}\perp\hspace{-0.95em}\perp}
\newcommand{\notindep}{\hspace{0.4em}\cancel{\perp\hspace{-0.95em}\perp}\hspace{0.4em}}
\newcommand{\T}{\mathrm{T}}
\newcommand{\diag}{\operatorname{diag}}
\renewcommand{\vec}{\operatorname{vec}}
\newcommand{\sbullet}{\,\mathbin{\vcenter{\hbox{\scalebox{.5}{$\bullet$}}}}\,}
\newcommand{\I}{\mathrm{I}}
\newcommand{\Val}{\mathit{Val}}
\newcommand{\dd}{\,d}
\newcommand{\HH}[1]{\operatorname{H}\left( #1 \right)}
\newcommand{\KL}[2]{\operatorname{KL}\left( #1 \,\middle\Vert\, #2 \right)}
\renewcommand{\O}[1]{O\left(#1\right)}
\newcommand{\X}{\mathbf{X}}
\newcommand{\x}{\mathbf{x}}
\newcommand{\G}{\mathcal{G}}
\newcommand{\B}{\mathcal{B}}
\newcommand{\D}{\mathcal{D}}
\newcommand{\PX}[1]{\Pi_{X_{#1}}}
\newcommand{\XP}[1]{X_{#1} \given \PX{#1}}
\newcommand{\TXi}{\Theta_{X_i}}
\newcommand{\TT}{\T \given \PX{i}}
\newcommand{\piijk}{\pi_{ik \given j}}
\newcommand{\dXi}{\delta_{X_i}}
\newcommand{\DXi}{\Delta_{X_i}}
\newcommand{\VDXi}{\Val(\DXi)}
\newcommand{\GXi}{\Gamma_{X_i}}
\newcommand{\idXi}{d_{X_i}}

\newcommand{\muB}{\boldsymbol{\mu}_{\B}}
\newcommand{\muBp}{\boldsymbol{\mu}_{\B'}}
\newcommand{\CB}{C_{\B}}
\newcommand{\CBp}{C_{\B'}}
\newcommand{\SB}{\Sigma_{\B}}
\newcommand{\SBp}{\Sigma_{\B'}}
\newcommand{\tSB}{\widetilde{\Sigma}_{\B}}
\newcommand{\tSBp}{\widetilde{\Sigma}_{\B'}}
\newcommand{\SmB}[1]{\Sigma_{#1}(\B)}
\newcommand{\bbX}[1]{\boldsymbol{\beta}_{X_{#1}}}
\newcommand{\eX}[1]{\varepsilon_{X_{#1}}}
\newcommand{\wx}[1]{\widehat{\mathbf{x}}_{#1}}
\newcommand{\we}[1]{\widehat{\boldsymbol{\varepsilon}}_{#1}}
\newcommand{\ws}[1]{\widehat{\sigma}^2_{#1}}
\newcommand{\wm}[1]{\widehat{\mu}_{#1}}
\newcommand{\wb}[1]{\widehat{\boldsymbol{\beta}}_{#1}}

\newcommand{\Db}[1]{\boldsymbol{\Delta}^{#1}}
\newcommand{\db}{\boldsymbol{\delta}}

\section{Introduction}
\label{sec:introduction}

Bayesian networks \citep[BNs; ][]{crc21} have played a central role in machine
learning research since the early days of the field as expert systems
\citep{castillo,cowell2}, graphical models \citep{pearl,koller}, dynamic
and latent variables models \citep{murphy}, and as the foundation of causal
discovery \citep{spirtes} and causal inference \citep{causality}. They have also
found applications as diverse as comorbidities in clinical psychology
\citep{mcnally}, the genetics of COVID-19 \citep{covid}, the Sustainable
Development Goals of the United Nations \citep{sdgs}, railway disruptions
\citep{railways} and industry 4.0 \citep{cirp}.

Machine learning, however, has evolved to include a variety of other models and
reformulated them into a very general information-theoretic framework. The
central quantities of this framework are Shannon's entropy and the
Kullback-Leibler divergence. Learning models from data relies crucially on the
former to measure the amount of information captured by the model (or its
complement, the amount of information lost in the residuals) and on the latter
as the loss function we want to minimise. For instance, we can construct
variational inference \citep{blei}, the Expectation-Maximisation algorithm
\cite{em}, Expectation Propagation \citep{ep} and various dimensionality
reduction approaches such as t-SNE \citep{tsne} and UMAP \citep{umap} using only
these two quantities. We can also reformulate classical maximum-likelihood and
Bayesian approaches to the same effect, from logistic regression to kernel
methods to boosting \citep{pml1,pml2}.

Therefore, the lack of literature on how to compute the entropy of a BN and the
Kullback-Leibler divergence between two BNs is surprising. While both are
mentioned in \citet{koller} and discussed at a theoretical level in
\citet{moral} for discrete BNs, no resources are available on any other type of
BN. Furthermore, no numerical examples of how to compute them are available even
for discrete BNs. \emph{We fill this gap in the literature by:}
\begin{itemize}
  \item \emph{Deriving efficient formulations of Shannon's entropy and the
    Kullback-Leibler divergence for Gaussian BNs and conditional
    linear Gaussian BNs.}
  \item \emph{Exploring the computational complexity of both for all common
    types of BNs.}
  \item \emph{Providing step-by-step numeric examples for all computations
    and all common types of BNs}.
\end{itemize}
Our aim is to make apparent how both quantities are computed in their
closed-form exact expressions and what is the associated computational cost.

The common alternative is to estimate both Shannon's entropy and the
Kullback-Leibler divergence empirically using Monte Carlo sampling. Admittedly,
this approach is simple to implement for all types of BNs. However, it has two
crucial drawbacks:
\begin{enumerate}
  \item Using asymptotic estimates voids the theoretical properties of many
    machine learning algorithms: Expectation-Maximisation is not guaranteed to
    converge \citep{koller}, for instance.
  \item The number of samples required to estimate the Kullback-Leibler
    divergence accurately on the tails of the global distribution of both BNs is
    also an issue \citep{gausmix}, especially when we need to evaluate it
    repeatedly as part of some machine learning algorithm. The same is true,
    although to a lesser extent, for Shannon's entropy as well. In general, the
    rate of convergence to the true posterior in Monte Carlo particle filters is
    proportional to the number of variables squared \citep{beskos}.
\end{enumerate}
Therefore, efficiently computing the exact value of Shannon's entropy and the
Kullback-Leibler divergence is a valuable research endeavour with a practical
impact on BN use in machine learning. To help its development, we implemented
the methods proposed in the paper in our \emph{bnlearn} R package
\citep{jss09}.

The remainder of the paper is structured as follows. In
Section~\ref{sec:definitions}, we provide the basic definitions, properties and
notation of BNs. In Section~\ref{sec:assumptions}, we revisit the most common
distributional assumptions in the BN literature: discrete BNs
(Section~\ref{sec:dbn}), Gaussian BNs (Section~\ref{sec:gbn}) and conditional
linear Gaussian BNs (Section~\ref{sec:cgbn}). We also briefly discuss exact and
approximate inference for these types of BNs in Section~\ref{sec:inference} to
introduce some key concepts for later use. In Section~\ref{sec:entropy-kl}, we
discuss how we can compute Shannon's entropy and the Kullback-Leibler divergence
for each type of BN. We conclude the paper by summarising and discussing the
relevance of these foundational results in Section~\ref{sec:conclusions}.
Appendix~\ref{app:bigO} summarises all the computational complexity results from
earlier sections, and Appendix~\ref{app:examples} contains additional examples
we omitted from the main text for brevity.

\section{Bayesian Networks}
\label{sec:definitions}

Bayesian networks (BNs) are a class of probabilistic graphical models defined
over a set of random variables $\X = \{X_1, \ldots, X_N\}$, each describing some
quantity of interest, that are associated with the nodes of a directed acyclic
graph (DAG) $\G$. Arcs in $\G$ express direct dependence relationships between
the variables in $\X$, with graphical separation in $\G$ implying conditional
independence in probability. As a result, $\G$ induces the factorisation
\begin{equation}
  \Prob{\X \given \G, \Theta} = \prod_{i=1}^N \Prob{\XP{i}, \TXi},
\label{eq:parents}
\end{equation}
in which the global distribution (of $\X$, with parameters $\Theta$) decomposes
into one local distribution for each $X_i$ (with parameters $\TXi$,
$\bigcup_{\X} \TXi = \Theta$) conditional on its parents $\PX{i}$.

This factorisation is as effective at reducing the computational
burden of working with BNs as the DAG underlying the BN is sparse, meaning that
each node $X_i$ has a small number of parents ($\lvert\PX{i}\rvert < c$, usually
with $c \in [2, 5]$). For instance, learning BNs from data is only feasible in
practice if this holds. The task of learning a BN $\B = (\G, \Theta)$ from a
data set $\D$ containing $n$ observations comprises two steps:
\begin{equation*}
  \underbrace{\Prob{\G, \Theta \given \D}}_{\text{learning}} =
    \underbrace{\Prob{\G \given \D}}_{\text{structure learning}} \cdot
    \underbrace{\Prob{\Theta \given \G, \D}}_{\text{parameter learning}}.
\end{equation*}
If we assume that parameters in different local distributions are independent
\citep{heckerman}, we can perform parameter learning independently for each
node. Each $\XP{i}$ will have a low-dimensional parameter space $\TXi$,
making parameter learning computationally efficient. On the other hand,
structure learning is well known to be both NP-hard \citep{nphard} and
NP-complete \citep{npcomp}, even under unrealistically favourable conditions
such as the availability of an independence and inference oracle
\citep{nplarge}. However, if $\G$ is sparse, heuristic learning algorithms have
been shown to run in quadratic time \citep{stco17}. Exact learning algorithms,
which have optimality guarantees that heuristic algorithms lack, retain their
exponential complexity but become feasible for small problems because sparsity
allows for tight bounds on goodness-of-fit scores and the efficient pruning of
the space of the DAGs \citep{cutting,suzuki17,scanagatta}.

\section{Common Distributional Assumptions for Bayesian Networks}
\label{sec:assumptions}

While there are many possible choices for the distribution of $\X$ in principle,
the literature has focused on three cases.

\subsection{Discrete BNs}
\label{sec:dbn}

\emph{Discrete BNs} \citep{heckerman} assume that both $\X$ and the $X_i$ are
multinomial random variables.\footnote{The literature sometimes denotes discrete
BNs as ``dBNs'' or ``DBNs''; we do not do that in this paper to avoid confusion
with dynamic BNs, which are also commonly denoted as ``dBNs''.} Local
distributions take the form
\begin{align*}
  &\XP{i} \sim \mathit{Mul}(\piijk),& &\piijk = \Prob{X_i = k \given \PX{i} = j};
\end{align*}
their parameters are the conditional probabilities of $X_i$ given each
configuration of the values of its parents, usually represented as a conditional
probability table (CPT) for each $X_i$. The $\piijk$ can be estimated from data
via the sufficient statistic $\{n_{ijk},$ $i = 1, \ldots N;$ $j = 1, \ldots,
q_i;$ $k = 1, \ldots, r_i\}$, the corresponding counts tallied from $\{X_i,
\PX{i}\}$ using maximum likelihood, Bayesian or shrinkage estimators as
described in \citet{koller} and \citet{shrinkent}.

The global distribution takes the form of an $N$-dimensional probability table
with one dimension for each variable. Assuming that each $X_i$ takes at most $l$
values, the table will contain $\lvert\Val(\X)\rvert = \O{l^N}$ cells, where
$\Val(\cdot)$ denotes the possible (configurations of the) values of its
argument. As a result, it is impractical to use for medium and large BNs.
Following standard practices from categorical data analysis \citep{agresti}, we
can produce the CPT for each $X_i$ from the global distribution by marginalising
(that is, summing over) all the variables other than $\{X_i, \PX{i}\}$ and then
normalising over each configuration of $\PX{i}$. Conversely, we can compose the
global distribution from the local distributions of the $X_i$ by multiplying the
appropriate set of conditional probabilities. The computational complexity of
the composition is $\O{Nl^N}$ because applying \mref{eq:parents} for each of the
$l^N$ cells yields
\begin{equation*}
  \Prob{\X = \x} = \prod_{i = 1}^N \Prob{X_i = x_i \given \PX{i} = \x_{\PX{i}}},
\end{equation*}
which involves $N$ multiplications. As for the decomposition, for each node, we:
\begin{enumerate}
  \item Sum over $N - |\PX{i}| - 1$ variables to produce the joint probability
    table for $\{X_i, \PX{i}\}$, which contains $\O{l^{|\PX{i}| + 1}}$ cells.
    The value of each cell is the sum of $\O{l^{N - |\PX{i}| - 1}}$
    probabilities.
  \item Normalise the columns of the joint probability table for
    $\{X_i, \PX{i}\}$ over each of the $\O{l^{|\PX{i}|}}$ configurations of
    values of $\PX{i}$,  which involves summing $\O{l}$ probabilities and
    dividing them by their total.
\end{enumerate}
The resulting computational complexity is
\begin{equation}
  \underbrace{
    \O{l^{|\PX{i}| + 1} \cdot l^{N - |\PX{i}| - 1}}
  }_{\text{marginalisation}} +
  \underbrace{
    \O{l \cdot l^{|\PX{i}|}}
  }_{\text{normalisation}} = \O{l^N + l^{|\PX{i}| + 1}}
\label{eq:marg-comp}
\end{equation}
for each node and $\O{Nl^N + l \sum_{i = 1}^N l^{|\PX{i}|}}$ for the whole BN.

\begin{Example}[Composing and decomposing a discrete BN]
  For reasons of space, this example is presented as Example~\ref{ex:dbn} in
  Appendix~\ref{app:examples}.
\label{ex:dbn-compose}
\end{Example}

\subsection{Gaussian BNs}
\label{sec:gbn}

\emph{Gaussian BNs} \citep[GBNs;][]{heckerman3} model $\X$ with a multivariate
normal random variable $N(\muB, \SB)$ and assume that the $X_i$ are univariate
normals linked by linear dependencies,
\begin{equation}
  \XP{i} \sim N(\mu_{X_i} + \PX{i}\bbX{i}, \sigma^2_{X_i}),
\label{eq:gbnlocal}
\end{equation}
which can be equivalently written as linear regression models of the form
\begin{align}
  &X_i = \mu_{X_i} + \PX{i}\bbX{i} + \eX{i},& &\eX{i} \sim N(0, \sigma^2_{X_i}).
\label{eq:gbnreg}
\end{align}
The parameters in \mref{eq:gbnlocal} and \mref{eq:gbnreg} are the regression
coefficients $\bbX{i}$ associated with the parents $\PX{i}$, an intercept term
$\mu_{X_i}$ and the variance $\sigma^2_{X_i}$. They are usually estimated by
maximum likelihood, but Bayesian and regularised estimators are available as
well \citep{crc21}.

The link between the parameterisation of the global distribution of a GBN and
that of its local distributions is detailed in \citet{pourahmadi}. We summarise
it here for later use.
\begin{itemize}
  \item \emph{Composing the global distribution.} We can create an $N \times N$
    lower triangular matrix $\CB$ from the regression coefficients in the local
    distributions such that $\CB\CB^\T$ gives $\SB$ after rearranging rows and
    columns. In particular, we:
    \begin{enumerate}
      \item Arrange the nodes of $\B$ in the (partial) topological ordering
        induced by $\G$, denoted $X_{(i)}, i = 1, \ldots, N$.
      \item The $i$th row of $\CB$ (denoted $\CB[i; \sbullet]$,
        $i = 1, \ldots, N)$ is associated with $X_{(i)}$. We compute its
        elements from the parameters of $\XP{(i)}$ as
        \begin{align*}
          &\CB[i; i] = \sqrt{\sigma^2_{X_{(i)}}}&
          &\text{and}&
          &\CB[i; \sbullet] = \bbX{(i)} \CB[\PX{(i)}; \sbullet],
        \end{align*}
        where $\CB[\PX{(i)}; \sbullet]$ are the rows of $\CB$ that
        correspond to the parents of $X_{(i)}$. The rows of $\CB$ are filled
        following the topological ordering of the BN.
      \item Compute $\tSB = \CB\CB^\T$.
      \item Rearrange the rows and columns of $\tSB$ to obtain $\SB$.
    \end{enumerate}
    Intuitively, we are constructing $\CB$ by propagating the node variances
    along the paths in $\G$ while combining them with the regression
    coefficients, which are functions of the correlations between adjacent
    nodes. As a result, $\CB\CB^\T$ gives $\SB$ after rearranging the rows and
    columns to follow the original ordering of the nodes.

    The elements of the mean vector $\muB$ are similarly computed as
    $\E(X_{(i)}) = \PX{(i)}\bbX{(i)}$ iterating over the
    variables in topological order.
  \item \emph{Decomposing the global distribution.} Conversely, we can derive
    the matrix $\CB$ from $\SB$ by reordering its rows and columns to follow the
    topological ordering of the variables in $\G$ and computing its Cholesky
    decomposition. Then
    \begin{equation*}
      R = \I_N - \diag(\CB)\CB^{-1},
    \end{equation*}
    contains the regression coefficients $\bbX{(i)}$ in the elements
    corresponding to $X_{(i)}, \PX{(i)}$.\footnote{Here $\diag(\CB)$ is a
    diagonal matrix with the same diagonal elements as $\CB$ and $\I_N$ is the
    identity matrix.} Finally, we compute the intercepts $\mu_{X_i}$ as
    $\muB - R \muB$ by reversing the equations we used to construct $\muB$
    above.

\end{itemize}
The computational complexity of composing the global distribution is bound by
the matrix multiplication $\CB\CB^\T$, which is $\O{N^3}$; if we assume that
$\G$ is sparse as in \citet{stco17}, the number of arcs is bound by some $cN$,
computing the $\muB$ takes $\O{N}$ operations. The complexity of decomposing the
global distribution is also $\O{N^3}$ because both inverting $\CB$ and
multiplying the result by $\diag(\CB)$ are $\O{N^3}$. \label{page:gbn-comp}

\begin{figure}[t]
  \centering
  \includegraphics[height=0.35\textheight]{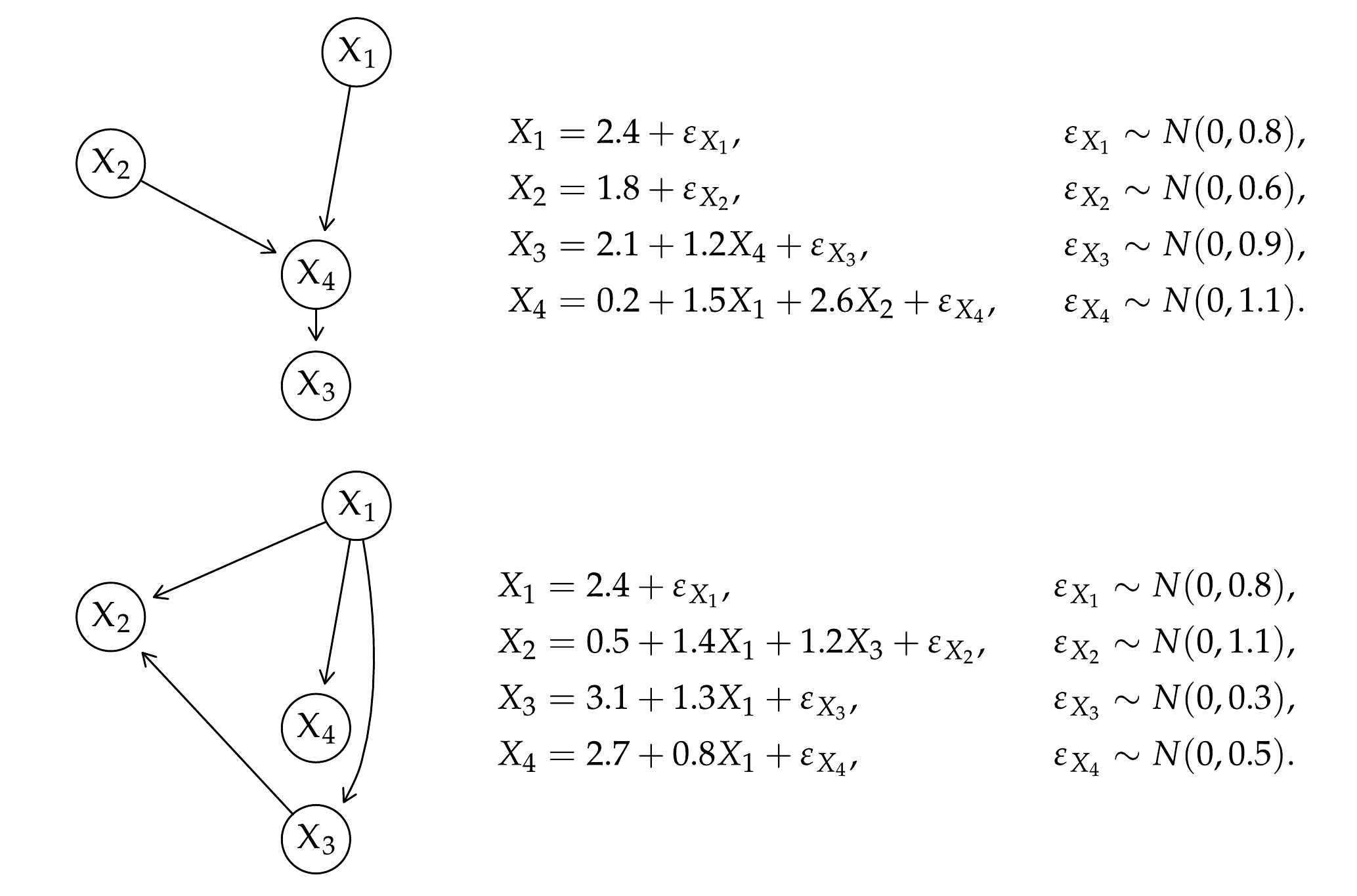}
  \caption{DAGs and local distributions for the GBNs $\B$ (top) and $\B'$
    (bottom) used in Examples~\ref{ex:gbn}, \ref{ex:gbn-h}, \ref{ex:mvnorm-kl},
    \ref{ex:gbn-kl} and~\ref{ex:gbn-approx-kl}.}
  \label{fig:gbns}
\end{figure}

\begin{Example}[Composing and decomposing a GBN]
  Consider the GBN $\B$ from Figure~\ref{fig:gbns} (top). The topological
  ordering of the variables defined by $\B$ is $\{\{X_1, X_2\}, X_4, X_3 \}$,
  so
  \begin{equation*}
    \CB = \bordermatrix{
        ~ & X_1    & X_2   & X_4  & X_3   \cr
      X_1 & 0.894 & 0     & 0     & 0     \cr
      X_2 & 0     & 0.774 & 0     & 0     \cr
      X_4 & 1.341 & 2.014 & 1.049 & 0     \cr
      X_3 & 1.610 & 2.416 & 1.258 & 0.948 \cr
    }
  \end{equation*}
  where the diagonal elements are
  \begin{align*}
    &\CB[X_1; X_1] = \sqrt{0.8},&
    &\CB[X_2; X_2] = \sqrt{0.6},&
    &\CB[X_4; X_4] = \sqrt{1.1},&
    &\CB[X_3; X_3] = \sqrt{0.9};
  \end{align*}
  and the elements below the diagonal are taken from the corresponding cells of
  \begin{align*}
    \CB[X_4; \sbullet] &=
      \begin{pmatrix} 1.5 & 2.6 \end{pmatrix}
      \begin{pmatrix} 0.894 & 0 & 0 & 0 \\
                      0 & 0.774 & 0 & 0 \end{pmatrix}, \\
    \CB[X_3; \sbullet] &=
      \begin{pmatrix} 1.2 \end{pmatrix}
      \begin{pmatrix} 1.341 & 2.014 & 1.049 & 0 \end{pmatrix}.
  \end{align*}
  Computing $\CB \CB^\T$ gives
  \begin{equation*}
    \tSB = \bordermatrix{
        ~ & X_1   & X_2   & X_4   & X_3    \cr
      X_1 & 0.800 & 0     & 1.200 &  1.440 \cr
      X_2 & 0     & 0.600 & 1.560 &  1.872 \cr
      X_4 & 1.200 & 1.560 & 6.956 &  8.347 \cr
      X_3 & 1.440 & 1.872 & 8.347 & 10.916 \cr
    }
  \end{equation*}
  and reordering the rows and columns of $\tSB$ gives
  \begin{equation*}
    \SB = \bordermatrix{
        ~ & X_1   & X_2   & X_3    & X_4   \cr
      X_1 & 0.800 & 0     &  1.440 & 1.200 \cr
      X_2 & 0     & 0.600 &  1.872 & 1.560 \cr
      X_3 & 1.440 & 1.872 & 10.916 & 8.347 \cr
      X_4 & 1.200 & 1.560 &  8.347 & 6.956 \cr
    }.
  \end{equation*}
  The elements of the corresponding expectation vector $\muB$ are then
  \begin{align*}
    \E(X_1) &= 2.400, \\
    \E(X_2) &= 1.800, \\
    \E(X_4) &= 0.2 + 1.5 \E(X_1) + 2.6 \E(X_2) = 8.480, \\
    \E(X_3) &= 2.1 + 1.2 \E(X_4) = 12.276.
  \end{align*}

  Starting from $\SB$, we can reorder its rows and columns to obtain $\tSB$. The
  Cholesky decomposition of $\tSB$ is $\CB$. Then
  \begin{align*}
    &\sigma^2_{X_1} = \CB[X_1; X_1]^2 = 0.8,&
    &\sigma^2_{X_2} = \CB[X_2; X_2]^2 = 0.6, \\
    &\sigma^2_{X_3} = \CB[X_3; X_3]^2 = 0.9,&
    &\sigma^2_{X_4} = \CB[X_4; X_4]^2 = 0.11.
  \end{align*}
  The coefficients $\bbX{i}$ of the local distributions are available from
  \begin{align*}
    R &= \I_N -
         \underbrace{\begin{bmatrix}
           0.894 & 0     & 0     & 0     \\
           0     & 0.774 & 0     & 0     \\
           0     & 0     & 1.049 & 0     \\
           0     & 0     & 0     & 0.948 \\
         \end{bmatrix}}_{\diag(\CB)}
         \underbrace{\begin{bmatrix}
           1.118 &  0     &  0     & 0     \\
           0     &  1.291 &  0     & 0     \\
          -1.430 & -2.479 &  0.953 & 0     \\
           0     &  0     & -1.265 & 1.054 \\
         \end{bmatrix}}_{\CB^{-1}} \\
      &= \bordermatrix{
        ~ & X_1   & X_2   & X_4   & X_3 \cr
      X_1 & 0     & 0     & 0     & 0   \cr
      X_2 & 0     & 0     & 0     & 0   \cr
      X_4 & 1.500 & 2.600 & 0     & 0   \cr
      X_3 & 0     & 0     & 1.200 & 0   \cr
    }
  \end{align*}
  where we can read
    $R_{X_4, X_1} = 1.5 = \beta_{X_4, X_1}$,
    $R_{X_4, X_2} = 2.6 = \beta_{X_4, X_2}$,
    $R_{X_3, X_4} = 1.2 = \beta_{X_3, X_4}$.

  We can read the standard errors of $X_1$, $X_2$, $X_3$ and $X_4$ directly from
  the diagonal elements of $\CB$, and we can compute the intercepts from
  $\muB - R \muB$ which amounts to
  \begin{align*}
    \mu_{X_1} &= \E(X_1) = 2.400, \\
    \mu_{X_2} &= \E(X_2) = 1.800, \\
    \mu_{X_4} &= \E(X_4) - \E(X_1)\beta_{X_4, X_1} -
                   \E(X_2)\beta_{X_4, X_2} = 0.200, \\
    \mu_{X_3} &= \E(X_3) - \E(X_4)\beta_{X_3, X_4} = 2.100.
  \end{align*}

\label{ex:gbn}
\end{Example}

\subsection{Conditional Linear Gaussian BNs}
\label{sec:cgbn}

Finally, \emph{conditional linear Gaussian BNs} \citep[CLGBNs;][]{lauritzen}
subsume discrete BNs and GBNs as particular cases by combining discrete and
continuous random variables in a mixture model. If we denote the former with
$\X_D$ and the latter with $\X_G$, so that $\X = \X_D \cup \X_G$, then:
\begin{itemize}
  \item Discrete $X_i \in \X_D$ are only allowed to have discrete parents
    (denoted $\DXi$), and are assumed to follow a multinomial distribution
    parameterised with CPTs. We can estimate their parameters in the same way as
    those in a discrete BN.
  \item Continuous $X_i \in \X_G$ are allowed to have both discrete and
    continuous parents (denoted $\GXi$, $\DXi \cup \GXi = \PX{i}$). Their
    local distributions are
    \begin{equation*}
      \XP{i} \sim N\left(\mu_{X_i, \dXi} +
                  \Gamma_{X_i}\boldsymbol{\beta}_{X_i, \dXi},
                    \sigma^2_{X_i, \dXi}\right),
    \end{equation*}
    which is equivalent to a mixture of linear regressions against the
    continuous parents with one component for each configuration $\dXi \in
    \VDXi$ of the discrete parents:
    \begin{align*}
      &X_i = \mu_{X_i, \dXi} + \Gamma_{X_i}\boldsymbol{\beta}_{X_i, \dXi} +
               \varepsilon_{X_i, \dXi},&
      &\varepsilon_{X_i, \dXi} \sim N\left(0, \sigma^2_{X_i, \dXi}\right).
    \end{align*}
    If $X_i$ has no discrete parents, the mixture reverts to a single linear
    regression like that in \mref{eq:gbnreg}. The parameters of these local
    distributions are usually estimated by maximum likelihood like those in a
    GBN; we have used hierarchical regressions with random effects in our recent
    work \citep{pgm22} for this purpose as well. Bayesian and regularised
    estimators are also an option \citep{koller}.
\end{itemize}
If the CLGBN comprises $\lvert \X_D \rvert = M$ discrete nodes and
$\lvert \X_G \rvert = N - M$ continuous nodes, these distributional assumptions
imply the partial topological ordering
\begin{equation}
  \underbrace{
    \left\{X_{(1)}, \ldots, X_{(M)} \right\}
  }_{\text{discrete nodes}}\, , \,
  \underbrace{
    \left\{X_{(M + 1)}, \ldots, X_{(N)} \right\}
  }_{\text{continuous nodes}}.
\label{eq:cgbn-partition}
\end{equation}
The discrete nodes jointly follow a multinomial distribution, effectively
forming a discrete BN. The continuous nodes jointly follow a multivariate normal
distribution, parameterised as a GBN, for each configuration of the discrete
nodes. Therefore, the global distribution is a Gaussian mixture in which the
discrete nodes identify the components, and the continuous nodes determine their
distribution. The practical link between the global and the local distributions
follows directly from Sections~\ref{sec:dbn} and~\ref{sec:gbn}.

\begin{Example}[Composing and decomposing a CLGBN]
  For reasons of space, this example is presented as Example~\ref{ex:cgbn} in
  Appendix~\ref{app:examples}.
\label{ex:cgbn-compose}
\end{Example}

The complexity of composing and decomposing the global
distribution is then
\begin{equation*}
  \underbrace{
    \O{M l^M}.
  }_{\text{convert between CPTs and component probabilities}} +
  \underbrace{
    \O{(N - M)^3 l^{\Val(\Db{})}}
  }_{\text{(de)compose the distinct component distributions}}
\end{equation*}
where $\Db{} = \bigcup_{X_i \in \X_G} \DXi$ are the discrete parents of the
continuous nodes. \label{page:cgbn-comp}

\subsection{Inference}
\label{sec:inference}

For BNs, \emph{inference} broadly denotes obtaining the conditional distribution
of a subset of variables conditional on a second subset of variables. Following
older terminology from expert systems \citep{castillo}, this is called
formulating a \emph{query} in which we ask the BN about the probability of an
\emph{event} of interest after observing some \emph{evidence}. In conditional
probability queries, the event of interest is the probability of one or more
events in (or the whole distribution of) some variables of interest conditional
on the values assumed by the evidence variables. In maximum a posteriori ("most
probable explanation") queries, we condition on the values of the evidence
variables to predict those of the event variables.

All inference computations on BNs are completely automated by \emph{exact} and
\emph{approximate} algorithms, which we will briefly describe here. We refer the
interested reader to the more detailed treatment in \citet{castillo} and
\citet{koller}.

Exact inference algorithms use local computations to compute the value of the
query. The seminal works of \citet{asia}, \citet{lauritzen} and
\citet{lauritzen2} describe how to transform a discrete BN or a (CL)GBN into a
\emph{junction tree}\footnote{A junction tree is an undirected tree whose nodes
are the cliques in the moral graph constructed from the BN and their
intersections. A clique a the maximal subset of nodes such that every two nodes
in the subset are adjacent.} as a preliminary step before using belief
propagation. \citet{cowell} uses elimination trees for the same purpose in
CLGBNs.

\citet{jtree-big0} give the computational complexity of constructing the
junction tree from a discrete BN as $\O{Nw + w l^w N}$ where $w$ is the maximum
number of nodes in a clique and, as before, $l$ is the maximum number of values
that a variable can take. We take the complexity of belief propagation to be
$\O{Nw l^w + \lvert\Theta\rvert}$, as stated in \citet{asia} (``The global
propagation is no worse than the initialisation [of the junction tree]''). This
is confirmed by \citet{pennock} and \citet{bp-big0}.

As for GBNs, we can also perform exact inference through their global
distribution because the latter has only $\O{N^2 + N}$ parameters. The
computational complexity of this approach is $\O{N^3}$ because of the cost of
composing the global distribution, which we derived in Section~\ref{sec:gbn}.
However, all the operations involved are linear, making it possible to leverage
specialised hardware such as GPUs and TPUs to the best effect.
\citet[][Section~14.2.1]{koller} note that ``inference in linear Gaussian
networks is linear in the number of cliques, and at most cubic in the size of
the largest clique'' when using junction trees and belief propagation.
Therefore, junction trees may be significantly faster for GBNs when $w \ll N$.
However, the correctness and convergence of belief propagation in GBNs require
a set of sufficient conditions that have been studied comprehensively by
\citet{malioutov}. Using the global distribution directly always produces
correct results.

Approximate inference algorithms use Monte Carlo simulations to sample from the
global distribution of $\X$ through the local distributions and estimate the
answer queries by computing the appropriate summary statistics on the particles
they generate. Therefore, they mirror the Monte Carlo and Markov chain Monte
Carlo approaches in the literature: rejection sampling, importance sampling, and
sequential Monte Carlo among others. Two state-of-the-art examples are the
\textit{adaptive importance sampling} (AIS-BN) scheme \citep{aisbn} and the
\textit{evidence pre-propagation importance sampling} (EPIS-BN) \citep{episbn}.

\section{Shannon Entropy and Kullback-Leibler Divergence}
\label{sec:entropy-kl}

The general definition of Shannon entropy for the probability distribution $P$
of $\X$ is
\begin{equation}
  \HH{P} = \E_{P}(-\log P(\X)) = - \int_{\Val(\X)} P(\x) \log P(\x) \dd\x.
\label{eq:entropy}
\end{equation}
The Kullback-Leibler divergence between two distributions $P$ and $Q$ for the
same random variables $\X$ is defined as
\begin{equation}
  \KL{P}{Q}
    = \E_{P(\X)}\left(-\log\frac{P(\X)}{Q(\X)}\right)
    = - \int_{\Val(\X)} \Prob{\x} \log\frac{P(\x)}{Q(\x)} \dd\x.
  \label{eq:kullback-leibler}
\end{equation}
They are linked as follows:
\begin{equation}
  \underbrace{
    \E_{P(\X)}\left(-\log\frac{P(\X)}{Q(\X)}\right)
  }_{\KL{P(\X)}{Q(\X)}} =
  \underbrace{
    \E_{P(\X)}(-\log P(\X))
  }_{\HH{P(\X)}} +
  \underbrace{
    \E_{P(\X)}\left(\log Q(\X)\right)
  }_{\HH{P(\X), Q(\X)}}
\label{eq:cross-entropy}
\end{equation}
where $\HH{P(\X), Q(\X)}$ is the cross-entropy between $P(\X)$ and $Q(\X)$. For
the many properties of these quantities, we refer the reader to \citet{itheory}
and \citet{csiszar}. Their use and interpretation are covered in depth (and
breadth!) in \citet{pml1,pml2} for general machine learning and in
\citet{koller} for BNs.

For a BN $\B$ encoding the probability distribution of $\X$, \mref{eq:entropy}
decomposes into
\begin{equation*}
  \HH{\B} = \sum_{i = 1}^N \HH{\XP{i}^{\B}}
\end{equation*}
where $\PX{i}^{\B}$ are the parents of $X_i$ in $\B$. While this decomposition
looks similar to \mref{eq:parents}, we will see that its terms are
not necessarily orthogonal unlike the local distributions.

As for \mref{eq:kullback-leibler}, we cannot simply write
\begin{equation*}
  \KL{\B}{\B'} = \sum_{i = 1}^N \KL{\XP{i}^{\B}}{\XP{i}^{\B'}}
\end{equation*}
because, in the general case, the nodes $X_i$ will have different parents in
$\B$ and $\B'$. This issue impacts the complexity of computing Kullback-Leibler
divergences in different ways depending on the type of BN.

\subsection{Discrete BNs}
\label{sec:dbn-kl}

For discrete BNs, $\HH{\B}$ does not decompose into orthogonal components. As
pointed out in \citet[][Section 8.4.12]{koller},
\begin{multline}
  \HH{\XP{i}^{\B}} =
    \sum_{j = 1}^{q_i} \Prob{\PX{i}^{\B} = j} \HH{\XP{i}^{\B} = j}
    \quad \text{where} \\
    \HH{\XP{i}^{\B} = j} =
      - \sum_{k = 1}^{r_i} \piijk(\B) \log \piijk(\B).
\label{eq:dbn-h}
\end{multline}
If we estimated the conditional probabilities $\piijk(\B)$ from data, the
$\Prob{\PX{i}^{\B} = j}$ are already available as the normalising constants of
the individual conditional distributions $\{\piijk(\B), j = 1, \ldots, q_i\}$ in
the local distribution of $X_i$. In this case, the complexity of computing
$\HH{\XP{i}^{\B}}$ is linear in the number of parameters:
$\O{\lvert\Theta\rvert} = \sum_{i = 1}^N \O{\lvert\TXi\rvert}$.

In the general case, we need exact inference to compute the probabilities
$\Prob{\PX{i}^{\B} = j}$. Fortunately, they can be readily extracted from the
junction tree derived from $\B$ as follows:
\begin{enumerate}
  \item Identify a clique containing both $X_i$ and $\PX{i}^{\B}$. Such a
    clique is guaranteed to exist by the family preservation property
    \citep[][Definition 10.1]{koller}.
  \item Compute the marginal distribution of $\PX{i}^{\B}$ by summing over
    the remaining variables in the clique.
\end{enumerate}
Combining the computational complexity of constructing the junction tree
from Section~\ref{sec:inference} and that of marginalisation, which is at most
$\O{l^{w - 1}}$ for each node as in \mref{eq:marg-comp}, we have
\begin{multline*}
  \underbrace{
    \O{Nw + w l^w N}
   }_{\text{create the junction tree}} +
   \underbrace{
     \O{Nl^{w - 1}}
   }_{\text{compute the $\Prob{\PX{i}^{\B} = j}$}} +
   \underbrace{
     \O{\lvert\Theta\rvert}
   }_{\text{compute $\HH{\B}$}} = \\
   \O{N(w(1 + l^w) + l^{w - 1}) + \lvert\Theta\rvert},
\end{multline*} \label{page:dbn-h-big0}
which is exponential in the maximum clique size $w$.\footnote{The maximum clique
size in a junction tree is proportional to the \emph{treewidth} of the BN the
junction tree is created from, which is also used in the literature to
characterise computational complexity in BNs.} Interestingly, we do not need to
perform belief propagation, so computing $\HH{\B}$ is more efficient than
other inference tasks.

\begin{figure}[t]
  \centering
  \includegraphics[height=0.35\textheight]{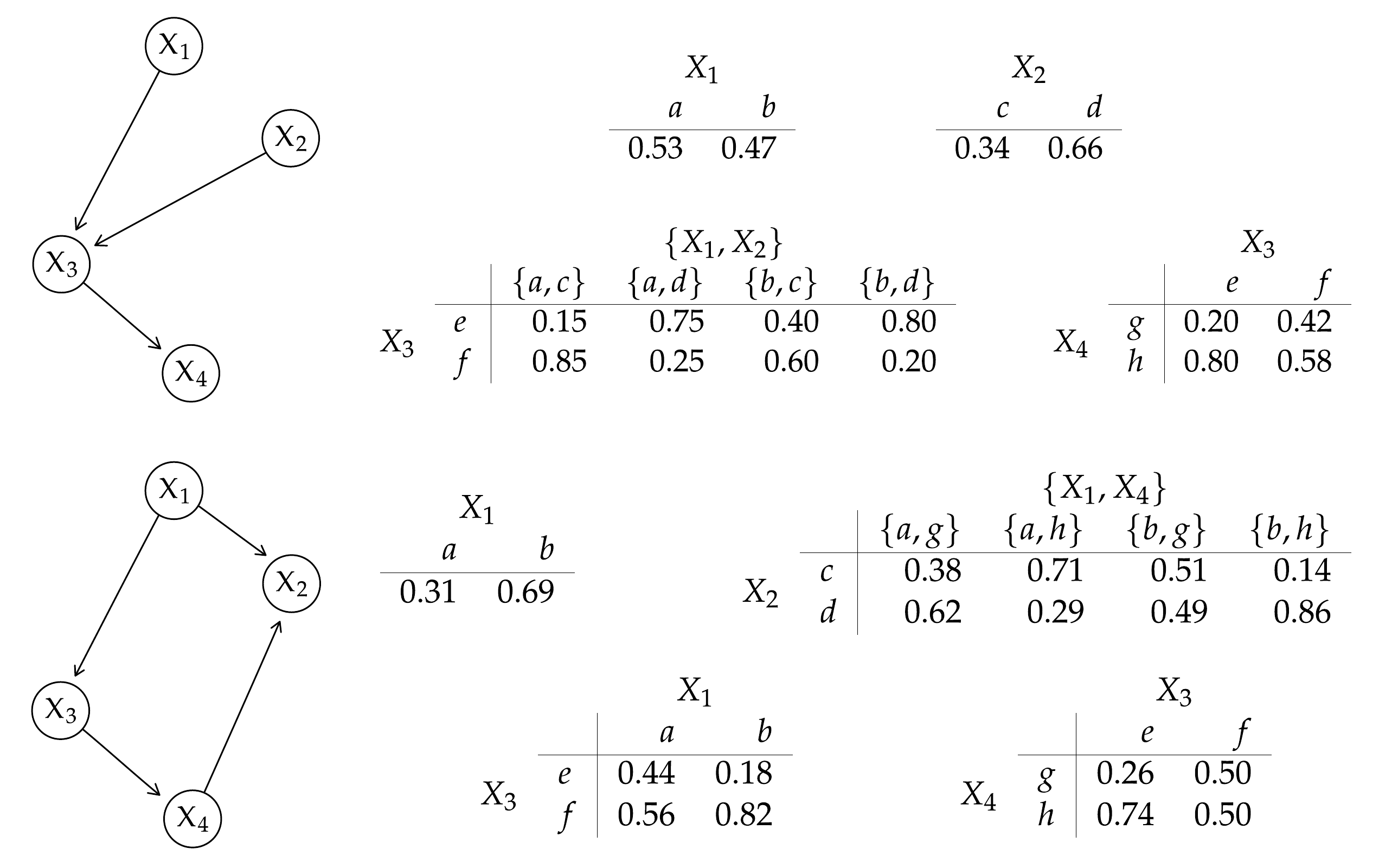}
  \caption{DAGs and local distributions for the discrete BNs $\B$ (top) and
    $\B'$ (bottom) used in Examples~\ref{ex:dbn-compose}, \ref{ex:dbn-h}
    and~\ref{ex:dbn-kl}.}
  \label{fig:dbns}
\end{figure}

\begin{Example}[Entropy of a discrete BN]
  For reasons of space, this example is presented as
  Example~\ref{ex:dbn-entropy} in Appendix~\ref{app:examples}.
\label{ex:dbn-h}
\end{Example}

The Kullback-Leibler divergence has a similar issue, as noted in
\citet[][Section 8.4.2]{koller}. The best and most complete explanation of how
to compute it for discrete BNs is in \citet{moral}. After decomposing
$\KL{\B}{\B'}$ following \mref{eq:cross-entropy} to separate $\HH{\B}$ and
$\HH{\B, \B'}$, \citet{moral} show that the latter takes the form
\begin{equation}
  \HH{\B, \B'} = \sum_{i = 1}^N
    \sum_{j \in \Val\left(\PX{i}^{\B'}\right)} \left[
    \sum_{k = 1}^{r_i}
    \pi_{ikj}(\B) \log \piijk(\B')
  \right]
\label{eq:dbn-hh}
\end{equation}
where:
\begin{itemize}
  \item $\pi_{ikj}(\B) = \Prob{X_i = k, \PX{i}(\B') = j}$ is the
    probability assigned by $\B$ to $X_i = k$ given that the variables that
    are parents of $X_i$ in $\B'$ take value $j$;
  \item $\piijk(\B') = \Prob{X_i = k \given \PX{i}(\B') = j}$ is the $(k, j)$
    element of the CPT of $X_i$ in $\B'$.
\end{itemize}
In order to compute the $\pi_{ikj}(\B)$, we need to transform $\B$ into its
junction tree and use belief propagation to compute the joint distribution of
$X_i \cup \PX{i}^{\B'}$. As a result, $\HH{\B, \B'}$ does not decompose at
all: each $\pi_{ikj}(\B)$ can potentially depend on the whole BN $\B$.

Algorithmically, to compute $\KL{\B}{\B'}$ we:
\begin{enumerate}
  \item Transform $\B$ into its junction tree.
  \item Compute the entropy $\HH{\B}$.
  \item For each node $X_i$:
    \begin{enumerate}
      \item Identify $\PX{i}^{\B'}$, the parents of $X_i$ in $\B'$.
      \item Obtain the distribution of the variables $\{ X_i, \PX{i}^{\B'} \}$
        from the junction tree of $\B$, consisting of the probabilities
        $\pi_{ikj}(\B)$.
      \item Read the $\piijk(\B')$ from the local
        distribution of $X_i$ in $\B'$.
    \end{enumerate}
  \item Use the $\pi_{ikj}(\B)$ and the $\piijk(\B')$ to compute
    \mref{eq:dbn-hh}.
\end{enumerate}
The computational complexity of this procedure is as follows:
\begin{multline}
   \underbrace{
     \O{N(w(1 + l^w) + l^{w - 1}) + \lvert\Theta\rvert}
   }_{\text{create the junction tree of $B$ and computing $\HH{\B}$}} +
   \underbrace{
     \O{N l^c (Nw l^w + \lvert\Theta\rvert)}
   }_{\text{produce the $\pi_{ikj}(\B)$}} +
   \underbrace{
     \O{\lvert\Theta\rvert}
   }_{\text{compute $\HH{\B, \B'}$}} = \\
   \O{N^2 w l^{w + c} + N(w + wl^w + l^{w - 1}) +
        (N l^c + 2) \lvert\Theta\rvert}.
\label{eq:dbn-kl-big0}
\end{multline}
As noted in \citet{moral}, computing the $\pi_{ikj}(\B)$ requires a separate run
of belief propagation for each configuration of the $\PX{i}^{\B'}$, for a total
of $\sum_{i = 1}^N l^{\lvert\PX{i}^{\B'}\rvert}$ times. If we assume that the
DAG underlying $\B'$ is sparse, we have that
$\lvert\PX{i}^{\B'}\rvert \leqslant c$ and the overall complexity of this step
becomes $\O{N l^c \cdot (Nw l^w + \lvert\Theta\rvert)}$, $N$ times that listed
in Section~\ref{sec:inference}. The caching scheme devised by \citet{moral} is
very effective in limiting the use of belief propagation, but it does not alter
its exponential complexity.

\begin{Example}[KL between two discrete BNs]
  Consider the discrete BN $\B$ from Figure~\ref{fig:dbns} (top). Furthermore,
  consider the BN $\B'$ from Figure~\ref{fig:dbns} (bottom). We constructed the
  global distribution of $\B$ in Example~\ref{ex:dbn}; we can similarly compose
  the global distribution of $\B'$, shown below.
  \begin{center}
    \small
    \setlength\extrarowheight{2.5pt}
    \begin{tabular}{r|rrp{0em}r|rrp{0em}r|rrp{0em}r|rr}
      \multicolumn{7}{c}{$X_1 = a$} & & \multicolumn{7}{c}{$X_1 = b$} \\
      \cline{1-7} \cline{9-15}
      \multicolumn{3}{c}{$X_2 = c$} & & \multicolumn{3}{c}{$X_2 = d$} & &
      \multicolumn{3}{c}{$X_2 = c$} & & \multicolumn{3}{c}{$X_2 = d$} \\
      \cline{1-3} \cline{5-7} \cline{9-11} \cline{13-15}
      \multicolumn{1}{c}{} & \multicolumn{2}{c}{$X_3$} & & \multicolumn{1}{c}{} & \multicolumn{2}{c}{$X_3$} & &
      \multicolumn{1}{c}{} & \multicolumn{2}{c}{$X_3$} & & \multicolumn{1}{c}{} & \multicolumn{2}{c}{$X_3$} \\
      $X_4$ & $e$ & $f$ & & $X_4$ & $e$ & $f$ & & $X_4$ & $e$ & $f$ & & $X_4$ & $e$ & $f$ \\
      \cline{1-3} \cline{5-7} \cline{9-11} \cline{13-15}
      $g$ & $0.013$ & $0.033$ & & $g$ & $0.022$ & $0.054$ & & $g$ & $0.016$ & $0.144$ & & $g$ & $0.016$ & $0.139$ \\
      $h$ & $0.072$ & $0.062$ & & $h$ & $0.029$ & $0.025$ & & $h$ & $0.013$ & $0.040$ & & $h$ & $0.079$ & $0.243$ \\
    \end{tabular}
  \end{center}

  \noindent Since both global distributions are limited in size, we can then
  compute the Kullback-Leibler divergence between $\B$ and $\B'$ using
  \mref{eq:kullback-leibler}.
  \begin{multline*}
    \KL{\B}{\B'} = - 0.013 \log 0.013 - 0.016 \log 0.016 - 0.022 \log 0.022
      - 0.016 \log 0.016 - \\ 0.072 \log 0.072 - 0.013 \log 0.013
      - 0.029 \log 0.029 - 0.079 \log 0.079 - \\ 0.033 \log 0.033
      - 0.144 \log 0.144 - 0.054 \log 0.054 - 0.139 \log 0.139 - \\
      - 0.062 \log 0.062 - 0.04 \log 0.04 - 0.025 \log 0.025 - 0.243 \log 0.243
      = 0.687
  \end{multline*}
  In the general case, when we cannot use the global distributions, we follow
  the approach described in Section~\ref{sec:dbn-kl}. Firstly, we apply
  \mref{eq:cross-entropy} to write
  \begin{equation*}
    \KL{\B}{\B'} = \HH{\B} - \HH{\B, \B'};
  \end{equation*}
  we have from Example~\ref{ex:dbn-entropy} that $\HH{\B} = 2.440$. As for the
  cross-entropy $\HH{\B, \B'}$, we apply \mref{eq:dbn-hh}:
  \begin{enumerate}
    \item We identify the parents of each node in $\B'$:
      \begin{align*}
        &\PX{1}^{\B'} = \{ \varnothing \},& &\PX{2}^{\B'} = \{ X_1, X_4\},&
        &\PX{3}^{\B'} = \{ X_1 \},& &\PX{4}^{\B'} = \{ X_3 \}.
      \end{align*}
    \item We construct a junction tree from $\B$ and we use it to compute the
      distributions $\Prob{X_1}$, $\Prob{X_2, X_1, X_4}$, $\Prob{X_3, X_1}$
      and $\Prob{X_4, X_3}$.
      \begin{center}
      \begin{tabular}{rr}
        \multicolumn{2}{c}{$X_1$} \\
        $a$ & $b$ \\
        \hline
        $0.53$ & $0.47$ \\
      \end{tabular}
      \hspace{0.10\linewidth}
      \begin{tabular}{ll|rrrr}
        \multicolumn{2}{c}{} & \multicolumn{4}{c}{$\{X_1, X_4\}$} \\
        & & $\{a, g\}$ & $\{a, h\}$ & $\{b, g\}$ & $\{b, h\}$ \\
        \cline{2-6}
        \multirow{2}{*}{$X_2$}
        & $c$ & $0.070$ & $0.110$ & $0.053$ & $0.107$ \\
        & $d$ & $0.089$ & $0.261$ & $0.076$ & $0.235$ \\
      \end{tabular}
      \end{center}
      \begin{center}
      \begin{tabular}{ll|rr}
        \multicolumn{2}{c}{} & \multicolumn{2}{c}{$X_1$} \\
        & & $a$ & $b$ \\
        \cline{2-4}
        \multirow{2}{*}{$X_3$}
        & $e$ & $0.289$ & $0.312$ \\
        & $f$ & $0.241$ & $0.158$ \\
      \end{tabular}
      \hspace{0.10\linewidth}
      \begin{tabular}{ll|rr}
        \multicolumn{2}{c}{} & \multicolumn{2}{c}{$X_3$} \\
        & & $e$ & $f$ \\
        \cline{2-4}
        \multirow{2}{*}{$X_4$}
        & $g$ & $0.120$ & $0.167$ \\
        & $h$ & $0.481$ & $0.231$ \\
      \end{tabular}
      \end{center}
    \item We compute the cross-entropy terms for the individual variables in
      $\B$ and $\B'$:
      \begin{align*}
        \HH{X_1^{\B}, X_1^{\B'}}
          &= 0.53 \log 0.31 + 0.47 \log 0.69 = -0.795; \\
        \HH{X_2^{\B}, X_2^{\B'}}
          &= 0.070 \log 0.38 + 0.089 \log 0.62 + 0.110 \log 0.71 +
             0.261 \log 0.29 + \\
          &\phantom{.=}   0.053 \log 0.51 + 0.076 \log 0.49 +
             0.107 \log 0.14 + 0.235 \log 0.86 \\
             &= -0.807; \\
        \HH{X_3^{\B}, X_3^{\B'}}
             &= 0.289 \log 0.44 + 0.241 \log 0.56 + 0.312 \log 0.18 +
                0.158 \log 0.82 \\
             &= -0.943; \\
        \HH{X_4^{\B}, X_4^{\B'}}
             &= 0.120 \log 0.26 + 0.481 \log 0.74 + 0.167 \log 0.50 +
                0.231 \log 0.50 \\
             &= -0.582;
      \end{align*}
      which sum up to $\HH{\B, \B'} = \sum_{i = 1}^N \HH{X_i^{\B}, X_i^{\B'}} =
      -3.127$.
    \item We compute $\KL{\B}{\B'} = 2.440 - 3.127 = 0.687$, which matches the
      value we previously computed from the global distributions.
  \end{enumerate}

\label{ex:dbn-kl}
\end{Example}

\subsection{Gaussian BNs}
\label{sec:gbn-kl}

$\HH{\B}$ decomposes along with the local distributions $\XP{i}$ in the case of
GBNs: from \mref{eq:gbnlocal}, each $\XP{i}$ is a univariate normal with
variance $\sigma^2_{X_i}(\B)$ and therefore
\begin{equation}
  \HH{\XP{i}^{\B}} =
    \frac{1}{2} \log\left( 2\pi\sigma^2_{X_i}(\B) \right) + \frac{1}{2}
\label{eq:gbn-h}
\end{equation} \label{page:gbn-h-big0}
which has a computational complexity of $\O{1}$ for each node, $\O{N}$ overall.
Equivalently, we can start from the global distribution of $\B$ from
Section~\ref{sec:gbn} and consider that
\begin{equation}
  \det(\Sigma) = \det(\CBp\CBp^\T) = \det(\CB)^2
    = \left(\prod\nolimits_{i = 1}^N \CB[i; i]\right)^2
    = \prod\nolimits_{i = 1}^N \sigma^2_{X_i}(\B)
\label{eq:klg-det}
\end{equation}
because $\CB$ is lower triangular. The (multivariate normal) entropy of $\X$
then becomes
\begin{multline*}
  \HH{\B}
    = \frac{N}{2} + \frac{N}{2} \log 2\pi + \frac{1}{2} \log \det(\Sigma)
    = \frac{N}{2} + \frac{N}{2} \log 2\pi + \frac{1}{2}
      \sum_{i = 1}^N \log \sigma^2_{X_i}(\B) \\
    = \sum_{i = 1}^N \frac{1}{2} +
      \frac{1}{2} \log\left(2\pi\sigma^2_{X_i}(\B)\right)
    = \sum_{i = 1}^N \HH{\XP{i}^{\B}}
\end{multline*}
in agreement with \mref{eq:gbn-h}.

\begin{Example}[Entropy of a GBN]
  For reasons of space, this example is presented as
  Example~\ref{ex:gbn-entropy} in Appendix~\ref{app:examples}.
\label{ex:gbn-h}
\end{Example}

In the literature, the Kullback-Leibler divergence between two GBNs $\B$ and
$\B'$ is usually computed using the respective global distributions
$N(\muB, \SB)$ and $N(\muBp, \SBp)$ \citep{gbnkl1,gbnkl2,gbnkl3}. The general
expression is
\begin{equation}
  \KL{\B}{\B'} =
    \frac{1}{2} \left[ \tr(\SBp^{-1} \SB) +
      (\muBp - \muB)^\T \SBp^{-1}(\muBp - \muB) -
      N + \log\frac{\det(\SBp)}{\det(\SB)}
    \right],
\label{eq:klg-global}
\end{equation}
which has computational complexity
\begin{multline}
  \underbrace{\O{2 N^3 + 2 N}}_{
    \text{compute $\muB$, $\muBp$ $\SB$, $\SBp$}
  } +
  \underbrace{\O{N^3}}_{
    \text{invert $\SBp$}
  } +
  \underbrace{\O{N^3}}_{
    \text{multiply $\SB^{-1}$ and $\SBp$}
  } +
  \underbrace{\O{N}}_{
    \text{trace of $\SB^{-1} \SBp$}
  } + \\
  \underbrace{\O{N^2 + 2N}}_{
    \text{compute $(\muBp - \muB)^\T \SBp^{-1}(\muBp - \muB)$}
  } +
  \underbrace{\O{N^3}}_{
    \text{determinant of $\SBp$}
  } +
  \underbrace{\O{N^3}}_{
    \text{determinant of $\SB$}
  } = \\
  \O{6N^3 + N^2 + 5N}.
\label{eq:klg-big0}
\end{multline}
The spectral decomposition $\SBp = U \Lambda_{\B'} U^\T$ gives the eigenvalues
$\diag(\Lambda_{\B'}) = \{\lambda_1(\B'), \ldots,$\linebreak$\lambda_N(\B') \}$
to compute $\SBp^{-1}$ and $\det(\SBp)$ efficiently as illustrated in the
example below. (Further computing the spectral decomposition of $\SB$ to compute
$\det(\SB)$ from the eigenvalues $\{\lambda_1(\B), \ldots, \lambda_N(\B) \}$
does not improve complexity because it just replaces a single $\O{N^3}$
operation with another one.) We thus somewhat improve the overall complexity of
$\KL{\B}{\B'}$ to $\O{5N^3 + N^2 + 6N}$.

\begin{Example}[General-case KL between two GBNs]
  Consider the GBN $\B$ Figure~\ref{fig:gbns} (top), which we know has global
  distribution
  \begin{align*}
    \begin{bmatrix} X_1 \\ X_2 \\ X_3 \\ X_4 \end{bmatrix} \sim
      N\left(
        \begin{bmatrix} 2.400 \\ 1.800 \\ 12.276 \\ 8.848 \end{bmatrix},
        \begin{bmatrix}
          0.800 & 0     &  1.440 & 1.200 \\
          0     & 0.600 &  1.872 & 1.560 \\
          1.440 & 1.872 & 10.916 & 8.347 \\
          1.200 & 1.560 &  8.347 & 6.956
        \end{bmatrix}
      \right)
  \end{align*}
  from Example~\ref{ex:gbn}. Furthermore, consider the GBN $\B'$ from
  Figure~\ref{fig:gbns} (bottom), which has global distribution
  \begin{align*}
    \begin{bmatrix} X_1 \\ X_2 \\ X_3 \\ X_4 \end{bmatrix} \sim
    N\left(
      \begin{bmatrix} 2.400 \\ 11.324 \\ 6.220 \\ 4.620 \end{bmatrix},
      \begin{bmatrix}
        0.800 & 2.368 & 1.040 & 0.640 \\
        2.368 & 8.541 & 3.438 & 1.894 \\
        1.040 & 3.438 & 1.652 & 0.832 \\
        0.640 & 1.894 & 0.832 & 1.012
      \end{bmatrix}
    \right).
  \end{align*}
  In order to compute $\KL{\B}{\B'}$, we first invert $\SBp$ to obtain
  \begin{equation*}
    \SBp^{-1} = \begin{bmatrix}
       9.945 & -1.272 & -2.806 & -1.600 \\
      -1.272 &  0.909 & -1.091 &  0      \\
      -2.806 & -1.091 &  4.642 &  0      \\
      -1.600 &  0     &  0     &  2.000
    \end{bmatrix},
  \end{equation*}
  which we then multiply by $\SB$ to compute the trace
  $\tr(\SBp^{-1}\SB) = 57.087$. We also use $\SBp^{-1}$ to compute
  $(\muBp - \muB)^\T \SBp^{-1}(\muBp - \muB) = 408.362$. Finally,
  $\det(\SBp) = 0.475$, $\det(\SB) = 0.132$ and therefore
  \begin{equation}
    \KL{\B}{\B'} =
      \frac{1}{2}\left[57.087 + 408.362 - 4 +
        \log\left(\frac{0.475}{0.132}\right)\right] = 230.0846.
  \label{eq:klg-ref}
  \end{equation}
  As an alternative, we can compute the spectral decompositions
  $\SB = U_\B \Lambda_\B U_\B^T$ and $\SBp = U_{\B'} \Lambda_{\B'} U_{\B'}^T$
  as an intermediate step. Multiplying the sets of eigenvalues
  \begin{align*}
    &\Lambda_\B = \diag(\{18.058, 0.741, 0.379, 0.093\})& &\text{and}&
    &\Lambda_{\B'} = \diag(\{11.106, 0.574, 0.236, 0.087\})
  \end{align*}
  gives the corresponding determinants; and it allow us to easily compute
  \begin{align*}
    &\SBp^{-1} = U_{\B'} \Lambda_{\B'}^{-1} U_{\B'}^T,& &\text{where}&
    &\Lambda_{\B'}^{-1} = \diag\left(\left\{\frac{1}{11.106}, \frac{1}{0.574},
                            \frac{1}{0.236}, \frac{1}{0.087}\right\}\right)
  \end{align*}
  for use in both the quadratic form and in the trace.
\label{ex:mvnorm-kl}
\end{Example}

However, computing $\KL{\B}{\B'}$ from the global distributions $N(\muB, \SB)$
and \linebreak $N(\muBp, \SBp)$ disregards the fact that BNs are sparse models
that can be characterised more compactly by $(\muB, \CB)$ and $(\muBp, \CBp)$ as
shown in Section~\ref{sec:gbn}. In particular, we can revisit several operations
that are in the high-order terms of \mref{eq:klg-big0}:
\begin{itemize}
  \item \emph{Composing the global distribution from the local ones.} We avoid
    computing $\SB$ and $\SBp$ thus reducing this step to $\O{2N}$ complexity.
  \item \emph{Computing the trace $\tr(\SBp^{-1} \SB)$.} We can reduce the
    computation of the trace as follows.
    \begin{enumerate}
      \item We can replace $\SB$ and $\SBp$ in the trace with any reordered
        matrix \citep[Result 8.17]{seber}: we choose to use $\tSBp$ and $\tSB^*$
        where $\tSBp$ is defined as before and $\tSB^*$ is $\SB$ with the rows
        and columns reordered to match $\tSBp$. Formally, this is equivalent to
        $\tSB^* = P \tSB P^\T$ where $P$ is a permutation matrix that imposes
        the desired node ordering: since both the rows and the columns are
        permuted in the same way, the diagonal elements of $\tSB$ are the same
        as those of $\tSB^*$ and the trace is unaffected.
      \item We have $\tSBp = \CBp\CBp^\T$.
      \item As for $\tSB^*$, we can write $\tSB^* = P \tSB P = (P \CB)
        (P \CB)^\T = \CB^* (\CB^*)^\T$ where \linebreak $\CB^* = P \CB$ is the
        lower triangular matrix $\CB$ with the rows re-ordered to match $\tSBp$.
        Note that $\CB^*$ is not lower triangular unless $\G$ and $\G'$ have the
        same partial node ordering, which implies $P = \I_N$.
    \end{enumerate}
    Therefore
    \begin{align}
      \tr(\SBp^{-1} \SB)
        = \tr\left((\CBp^{-1}\CB^*)^\T(\CBp^{-1}\CB^*)\right)
        = \| \CBp^{-1}\CB^* \|_F^2
      \label{eq:klg-tr}
    \end{align}
    where the last step rests on \citet[Result 4.15]{seber}. We can invert
    $\CBp$ in $\O{N^2}$ time following \citet[Algorithm 2.3]{stewart}.
    Multiplying $\CBp^{-1}$ and $\CB^*$ is still $\O{N^3}$. The Frobenius norm
    $\|\cdot\|_F$ is $\O{N^2}$ since it's the sum of the squared elements of
    $\CBp^{-1}\CB^*$.
  \item \emph{Computing the determinants $\det(\SBp)$ and $\det(\SB)$.}
    From \mref{eq:klg-det}, each determinant can be computed in $O(N)$.
  \item \emph{Computing the quadratic term $(\muBp - \muB)^\T \SBp^{-1}
    (\muBp - \muB)$}. Decomposing $\SBp^{-1}$ leads to
    \begin{equation}
      (\muBp - \muB)^\T \SBp^{-1}(\muBp - \muB) =
      (\CBp^{-1}(\muBp^* - \muB^*))^\T \CBp^{-1}(\muBp^* - \muB^*),
    \label{eq:klg-quad}
    \end{equation}
    where $\muBp^*$ and $\muB^*$ are the mean vectors re-ordered to match
    $\CBp^{-1}$. The computational complexity is still $\O{N^2 + 2N}$ because
    $\CBp^{-1}$ is available from previous computations.
\end{itemize}

\noindent Combining \mref{eq:klg-tr}, \mref{eq:klg-det} and \mref{eq:klg-quad},
the expression in \mref{eq:klg-global} becomes
\begin{multline}
  \KL{\B}{\B'} = \\
    \frac{1}{2} \left[ \| \CBp^{-1}\CB^* \|_F^2 +
    (\CBp^{-1}(\muBp^* - \muB^*))^\T \CBp^{-1}(\muBp^* - \muB^*) -
      N + 2\log\frac{\prod\nolimits_{i = 1}^N \CBp[i; i]}
                    {\prod\nolimits_{i = 1}^N \CB[i; i]}
    \right].
\label{eq:klg-faster}
\end{multline}
The overall complexity of \mref{eq:klg-faster} KL is
\begin{multline}
  \underbrace{\O{2 N^2 + 2 N}}_{
    \text{compute $\muB$, $\muBp$ $\CB$, $\CBp$}
  } +
  \underbrace{\O{2N^2 + N^3}}_{
    \text{compute $\| \CBp^{-1}C_{\B} \|_F^2$}
  } +
  \underbrace{\O{N^2 + 2N}}_{
    \text{compute the quadratic form}
  } + \\
  \underbrace{\O{2N}}_{
    \text{compute $\det(\SB)$, $\det(\SBp)$}
  } =
  \O{N^3 + 5N^2 + 6N};
\label{eq:klg-fast0}
\end{multline}
while still cubic, the leading coefficient suggests that it should be about
5 times faster than the variant of \mref{eq:klg-big0} using the spectral
decomposition.

\begin{Example}[Sparse KL between two GBNs]
  Consider again the two GBNs from Example~\ref{ex:mvnorm-kl}. The corresponding
  matrices
  \begin{align*}
    &\CB = \bordermatrix{
        ~ & X_1   & X_2   & X_4   & X_3   \cr
      X_1 & 0.894 & 0     & 0     & 0     \cr
      X_2 & 0     & 0.774 & 0     & 0     \cr
      X_4 & 1.341 & 2.014 & 1.049 & 0     \cr
      X_3 & 1.610 & 2.416 & 1.258 & 0.948 \cr
    },&
    &\CBp = \bordermatrix{
        ~ & X_1   & X_3   & X_4   & X_2   \cr
      X_1 & 0.894 & 0     & 0     & 0     \cr
      X_3 & 1.163 & 0.548 & 0     & 0     \cr
      X_4 & 0.715 & 0     & 0.707 & 0     \cr
      X_2 & 2.647 & 0.657 & 0     & 1.049 \cr
     }
  \end{align*}
  readily give the determinants of $\SB$ and $\SBp$ following \mref{eq:klg-det}:
  \begin{align*}
    \det(\CB) &= (0.894 \cdot 0.774 \cdot 1.049 \cdot 0.948)^2 = 0.475, \\
    \det(\CBp) &= (0.894 \cdot 0.548 \cdot 0.707 \cdot 1.049)^2 = 0.132.
  \end{align*}
  As for the Frobenius norm in \mref{eq:klg-tr}, we first invert $\CBp$ to
  obtain
  \begin{equation*}
    \CBp^{-1} = \bordermatrix{
        ~ &  X_1   &  X_3   & X_4   & X_2   \cr
      X_1 &  1.118 &  0     & 0     & 0     \cr
      X_3 & -2.373 &  1.825 & 0     & 0     \cr
      X_4 & -1.131 &  0     & 1.414 & 0     \cr
      X_2 & -1.334 & -1.144 & 0     & 0.953 \cr
    };
  \end{equation*}
  then we reorder the rows and columns of $\CB$ to follow the same node
  ordering as $\CBp$ and compute
  \begin{equation*}
    \left\|\begin{pmatrix}
       1.118 &  0     & 0     & 0     \\
      -2.373 &  1.825 & 0     & 0     \\
      -1.131 &  0     & 1.414 & 0     \\
      -1.334 & -1.144 & 0     & 0.953 \\
    \end{pmatrix}\begin{pmatrix}
      0.894 & 0     & 0     & 0     \\
      1.610 & 0.948 & 1.258 & 2.416 \\
      1.341 & 0     & 1.049 & 2.014 \\
      0     & 0     & 0     & 0.774 \\
    \end{pmatrix}\right\|_F^2 = 57.087
  \end{equation*}
  which, as expected, matches the value of $\tr(\SBp^{-1}\SB)$ we computed in
  Example~\ref{ex:mvnorm-kl}. Finally, \linebreak $\CBp^{-1}(\muBp^* - \muB^*)$
  in \mref{eq:klg-quad} is
  \begin{equation*}
      \begin{pmatrix}
       1.118 &  0     & 0     & 0     \\
      -2.373 &  1.825 & 0     & 0     \\
      -1.131 &  0     & 1.414 & 0     \\
      -1.334 & -1.144 & 0     & 0.953 \\
    \end{pmatrix}
    \left[
      \begin{pmatrix}
        2.400  \\
        6.220  \\
        4.620  \\
        11.324 \\
      \end{pmatrix} -
      \begin{pmatrix}
        2.400   \\
        12.1276 \\
        8.848   \\
        1.800   \\
      \end{pmatrix}
    \right] = \begin{pmatrix}
          0     \\
        -11.056 \\
         -5.459 \\
         16.010 \\
      \end{pmatrix}.
  \end{equation*}
  The quadratic form is then equal to $408.362$, which matches the value of
  \mbox{$(\muBp - \muB)^\T \SBp^{-1}(\muBp - \muB)$} in
  Example~\ref{ex:mvnorm-kl}. As a result, the expression for $\KL{\B}{\B'}$ is
  the same as in \mref{eq:klg-ref}.
\label{ex:gbn-kl}
\end{Example}

We can further reduce the complexity \mref{eq:klg-fast0} of \mref{eq:klg-faster}
when an approximate value of KL is suitable for our purposes. The only term with
cubic complexity is \mbox{$\tr(\SBp^{-1} \SB) = \|\CBp^{-1}\CB^* \|_F^2$}:
reducing it to quadratic complexity or lower will eliminate the leading term of
\mref{eq:klg-fast0}, making it quadratic in complexity. One way to do this is to
compute a lower and an upper bound for $\tr(\SBp^{-1} \SB)$, which can serve as
an interval estimate, and take their geometric mean as an approximate point
estimate.

A lower bound is given by \citet[Result 10.39]{seber}:
\begin{equation}
  \tr(\SBp^{-1} \SB) \geqslant \log\det(\SBp^{-1} \SB) + N =
    -\log\det(\SBp) + \log\det(\SB) + N,
\label{eq:trace-lb}
\end{equation}
which conveniently reuses the values of $\det(\SB)$ and $\det(\SBp)$ we have
from \mref{eq:klg-det}. For an upper bound, \citet[Result 10.59]{seber}
combined with \citet[Result 4.15]{seber} gives
\begin{equation}
  \tr(\SBp^{-1} \SB) \leqslant \tr(\SBp^{-1})\tr(\SB) =
    \tr\left((\CBp\CBp^\T)^{-1}\right)\tr(\CB\CB^\T) =
    \|\CBp^{-1}\|^2_F \|\CB\|^2_F \, ,
\label{eq:trace-ub}
\end{equation}
a function of $\CB$ and $\CBp$ that can be computed in $\O{2N^2}$ time. Note
that, as far as the point estimate is concerned, we do not care about how wide
the interval is: we only need its geometric mean to be an acceptable
approximation of $\tr(\SBp^{-1} \SB)$.

\begin{Example}[Approximate KL]
  From Example~\ref{ex:mvnorm-kl}, we have that $\tr(\SBp^{-1}\SB) = 57.087$,
  \linebreak \mbox{$\det(\SBp) = 0.475$} and $\det(\SB) = 0.132$. The lower
  bound in \mref{eq:trace-lb} is then
  \begin{equation*}
    -\log\det(\SBp) + \log\det(\SB) + 4 = 5.281
  \end{equation*}
  and the upper bound in \mref{eq:trace-ub} is
  \begin{equation*}
    \|\CBp^{-1}\|^2_F \|\CB\|^2_F = 17.496 \cdot 19.272 = 337.207.
  \end{equation*}
  Their geometric mean is $42.199$, which can serve as an approximate value for
  $\KL{\B}{\B'}$.
\label{ex:gbn-approx-kl}
\end{Example}

If we are comparing two GBNs whose parameters (but not necessarily network
structures) have been learned from the same data, we can sometimes approximate
$\KL{\B}{\B'}$ using the local distributions $\XP{i}^{\B}$ and $\XP{i}^{\B'}$
directly. If $\B$ and $\B'$ have compatible partial orderings,\footnote{By
``compatible partial orderings'', we mean two partial orderings that can be
sorted into at least one shared total node ordering that is compatible with
both.} we can define a common total node ordering for both such that
\begin{align*}
  \KL{\B}{\B'}
    &= \KL{X_{(1)} \given \{X_{(2)}, \ldots, X_{(N)}\} \cdots X_{N}}
          {X_{(1)} \given \{X_{(2)}, \ldots, X_{(N)}\} \cdots X_{N}} \\
    &= \KL{\XP{(1)}^{\B} \cdot \ldots \cdot \XP{(N)}^{\B}}
          {\XP{(1)}^{\B'} \cdot \ldots \cdot \XP{(N)}^{\B'}}.
\end{align*}
The product of the local distributions in the second step is obtained from
the chain decomposition in the first step by considering the nodes in the
conditioning other than the parents to have associated regression coefficients
equal to zero. Then, following the derivations in \citet{cavanaugh} for a
general linear regression model, we can write the empirical approximation
\begin{equation}
  \KL{\XP{i}^{\B}}{\XP{i}^{\B'}} \approx
    \frac{1}{2} \left( \log\frac{\ws{X_i}(\B')}{\ws{X_i}(\B)}
      + \frac{\ws{X_i}(\B)}{\ws{X_i}(\B')} - 1\right)
    + \frac{1}{2n} \left(
        \frac{\| \wx{i}(\B) - \wx{i}(\B') \|^2_2}{\ws{X_i}(\B')} \right)
\label{eq:klg-lrm}
\end{equation}
where, following a similar notation to \mref{eq:gbnreg}:
\begin{itemize}
  \item $\wm{X_i}(\B)$, $\wb{X_i}(\B)$, $\wm{X_i}(\B')$, $\wb{X_i}(\B')$ are the
    estimated intercepts and regression coefficients;
  \item $\wx{i}(\B)$ and $\wx{i}(\B')$ are the $n \times 1$ vectors
    \begin{align*}
      &\wx{i}(\B) = \wm{X_i}(\B) + \x[\sbullet ; \PX{i}(\B)] \wb{X_i}(\B),&
      &\wx{i}(\B') = \wm{X_i}(\B') + \x[\sbullet ; \PX{i}(\B')] \wb{X_i}(\B'),
    \end{align*}
    the fitted values computed from the data observed for $X_i$,
    $\PX{i}(\B)$, $\PX{i}(\B')$;
  \item $\sigma^2_{X_i}(\B)$ and $\sigma^2_{X_i}(\B')$ are the residual
    variances in $\B$ and $\B'$.
\end{itemize}
We can compute the expression in \mref{eq:klg-lrm} for each node in
\begin{multline*}
  \underbrace{\O{n(\lvert \PX{i}(\B) \rvert + \lvert \PX{i}(\B') \rvert + 2)}}_{
    \text{compute $\wx{i}(\B)$ and $\wx{i}(\B')$}
  } +
  \underbrace{\O{n}}_{
    \text{compute the norm $\| \wx{i}(\B) - \wx{i}(\B') \|^2_2$}
  } = \\
  \O{n(\lvert \PX{i}(\B) \rvert + \lvert \PX{i}(\B') \rvert + \sfrac{5}{2})},
\end{multline*} \label{page:gbn-kl-lrm-big0}
which is linear in the sample size if both $\G$ and $\G'$ are
sparse because $\lvert \PX{i}(\B) \rvert \leqslant c$, $\lvert \PX{i}(\B')
\rvert \leqslant c$. In this case, the overall computational complexity
simplifies to \linebreak $O(nN(2c + \sfrac{5}{2}))$. Furthermore, as we pointed
out in \citet{stco17}, the fitted values $\wx{i}(\B)$, $\wx{i}(\B')$ are
computed as a by-product of parameter learning: if we consider them to be
already available, the above computational complexity reduced to just $\O{n}$
for a single node and $\O{nN}$ overall. We can also replace the fitted values
$\wx{i}(\B)$, $\wx{i}(\B')$ in \mref{eq:klg-lrm} with the corresponding
residuals $\we{i}(\B)$, $\we{i}(\B')$ because
\begin{multline*}
  \| \wx{i}(\B) - \wx{i}(\B') \|_2^2 =
    \| (\x[\sbullet; X_i] - \wx{i}(\B)) -
       (\x[\sbullet; X_i] - \wx{i}(\B')) \|_2^2 =
    \|\we{i}(\B) - \we{i}(\B') \|_2^2
\end{multline*}
if the latter are available but the former are not.

\begin{Example}[KL between GBNs with parameters estimated from data]
  For reasons of space, this example is presented as Example~\ref{ex:gbn-lrm}
  in Appendix~\ref{app:examples}.
\end{Example}

\subsection{Conditional Gaussian BNs}
\label{sec:cgbn-kl}

\begin{figure}[t]
  \centering
  \includegraphics[height=0.55\textheight]{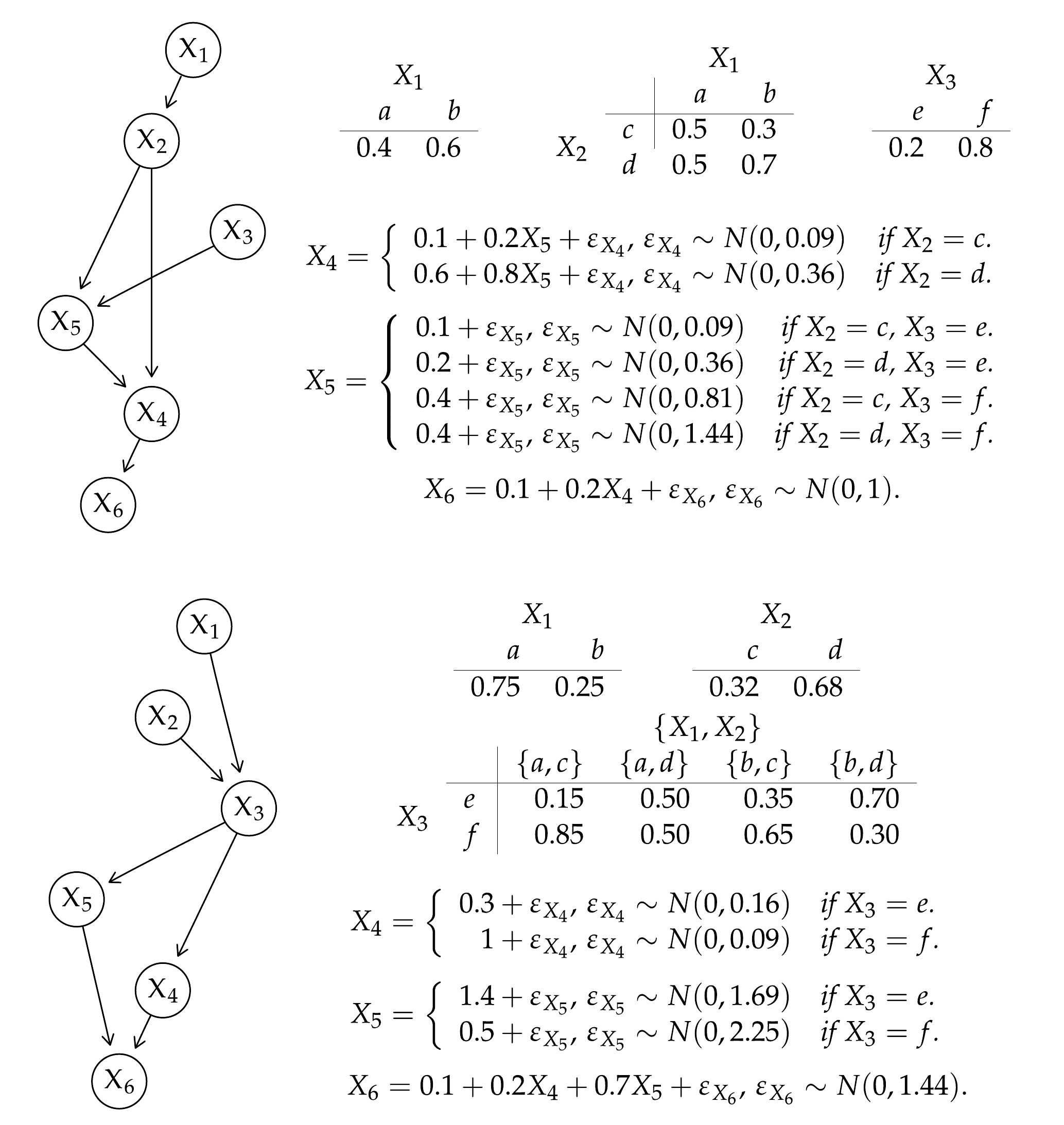}
  \caption{DAGs and local distributions for the CLGBNs $\B$ (top) and $\B'$
    (bottom) used in Examples~\ref{ex:cgbn-compose}, \ref{ex:cgbn-h},
    \ref{ex:cgbn-kl} and~\ref{ex:cgbn-kl-sparse}.}
  \label{fig:cgbns}
\end{figure}

The entropy $\HH{\B}$ decomposes into a separate $\HH{\XP{i}^{\B}}$ for each
node, of the form \mref{eq:dbn-h} for discrete nodes and \mref{eq:gbn-h} for
continuous nodes with no discrete parents. For continuous nodes with both
discrete and continuous parents,
\begin{equation}
  \HH{\XP{i}^{\B}} = \frac{1}{2} \sum_{\dXi \in \VDXi} \pi_{\dXi}
    \log\left(2\pi\sigma^2_{X_i, \dXi}(\B)\right) + \frac{1}{2},
\label{eq:cgbn-h}
\end{equation}
where $\pi_{\dXi}$ represents the probability associated with the configuration
$\dXi$ of the discrete parents $\DXi$. This last expression can be computed in
$\O{\lvert\VDXi\rvert}$ time for each node. Overall, the complexity of computing
$\HH{\B}$ is
\begin{equation*}
  \O{\sum_{X_i \in \X_{D}} \lvert \TXi \rvert +
     \sum_{X_i \in \X_{G}} \max\left\{1, \lvert\VDXi\rvert \right\}}.
\end{equation*} \label{page:cgbn-h-big0}
where the $\max$ accounts for the fact that $\lvert\VDXi\rvert = 0$ when
$\DXi = \varnothing$ but the computational complexity is $\O{1}$ for such nodes.

\begin{Example}[Entropy of a CLGBN]
  For reasons of space, this example is presented as
  Example~\ref{ex:cgbn-entropy} in Appendix~\ref{app:examples}.
\label{ex:cgbn-h}
\end{Example}

As for $\KL{\B}{\B'}$, we could not find any literature illustrating how to
compute it. The partition of the nodes in \mref{eq:cgbn-partition} implies that
\begin{equation}
  \KL{\B}{\B'} =
    \underbrace{
      \KL{\X_{D}^{\B}}{\X_{D}^{\B'}}
    }_{\text{discrete nodes}} +
    \underbrace{
      \KL{\X_{G}^{\B} \given \X_{D}^{\B}}{\X_{G}^{\B'} \given \X_{D}^{\B'}}
    }_{\text{continuous nodes}}.
\label{eq:cgbn-kl}
\end{equation}
We can compute the first term following Section~\ref{sec:dbn-kl}:
$\X_{D}^{\B}$ and $\X_{D}^{\B'}$ form two discrete BNs whose DAGs are the
spanning subgraphs of $\B$ and $\B'$ and whose local distributions are the
corresponding ones in $\B$ and $\B'$, respectively. The second term decomposes
into
\begin{equation}
  \KL{\X_{G}^{\B} \given \X_{D}^{\B}}{\X_{G}^{\B'} \given \X_{D}^{\B'}} =
    \sum_{\x_{D} \in \Val(\X_D)} \Prob{\X_{D}^{\B} = \x_{D}}
    \KL{\X_{G}^{\B} \given \X_{D}^{\B} = \x_{D}}
       {\X_{G}^{\B'} \given \X_{D}^{\B'} = \x_{D}}
\label{eq:cgbn-kl-2nd}
\end{equation}
similarly to \mref{eq:dbn-hh} and \mref{eq:cgbn-h}. We can compute it using the
multivariate normal distributions associated with the $\X_{D}^{\B} = \x_{D}$
and the $\X_{D}^{\B'} = \x_{D}$ in the global distributions of $\B$ and $\B'$.

\begin{Example}[General-case KL between two CLGBNs]
  Consider the CLGBNs $\B$ from Figure~\ref{fig:cgbns} (top), which we already
  used in Examples~\ref{ex:cgbn-compose} and~\ref{ex:cgbn-h}, and $\B'$ from
  Figure~\ref{fig:cgbns} (bottom). The variables $\X_{D}^{\B'}$ identify the
  following mixture components in the global distribution of $\B'$:
  \begin{align*}
    \{a, c, e\}, \{b, c, e\}, \{a, d, e\}, \{b, d, e\} &\mapsto \{e\}, \\
    \{a, c, f\}, \{b, c, f\}, \{a, d, f\}, \{b, d, f\} &\mapsto \{f\}.
  \end{align*}
  Therefore, $\B'$ only encodes two different multivariate normal distributions.

  Firstly, we construct two discrete BNs using the subgraphs spanning
  $\X_{D}^{\B} = \X_{D}^{\B'} = \{X_1, X_2, X_3\}$ in $\B$ and
  $\B'$, which have arcs $\{X_1 \rarr X_2\}$ and $\{X_1 \rarr X_2, X_2 \rarr
  X_3 \}$ respectively.
  The CPTs for $X_1$, $X_2$ and $X_3$ are the same as in $\B$ and in $\B'$. We
  then compute $\KL{\X_{D}^{\B}}{\X_{D}^{\B'}} = 0.577$ following
  Example~\ref{ex:dbn-kl}.

  Secondly, we construct the multivariate normal distributions associated with
  the components of $\B'$ following Example~\ref{ex:cgbn-compose} (in which we
  computed those of $\B$). For $\{e\}$, we have
  \begin{align*}
    \begin{bmatrix} X_4 \\ X_5 \\ X_6 \end{bmatrix} \sim
      N\left(
        \begin{bmatrix} 0.300 \\ 1.400 \\ 1.140 \end{bmatrix},
        \begin{bmatrix}
          0.160 & 0.000 & 0.032 \\
          0.000 & 1.690 & 1.183 \\
          0.032 & 1.183 & 2.274
        \end{bmatrix}
      \right);
  \end{align*}
  for $\{f\}$, we have
  \begin{align*}
    \begin{bmatrix} X_4 \\ X_5 \\ X_6 \end{bmatrix} \sim
      N\left(
        \begin{bmatrix} 1.000 \\ 0.500 \\ 0.650 \end{bmatrix},
        \begin{bmatrix}
          0.090 & 0.000 & 0.018 \\
          0.000 & 2.250 & 1.575 \\
          0.018 & 1.575 & 2.546
        \end{bmatrix}
      \right).
  \end{align*}
  Then,
  \begin{align*}
    &\KL{\X_{G}^{\B} \given \X_{D}^{\B}}{\X_{G}^{\B'} \given \X_{D}^{\B'}} \\
    &= \sum_{x_1 \in \{a, b\}} \sum_{x_2 \in \{c, d\}} \sum_{x_3 \in \{e, f\}}
         \Prob{\X_{D}^{\B} = \{x_1, x_2, x_3\}} \,\cdot\\
    &\hspace{4em} \KL{\X_{G}^{\B} \given \X_{D}^{\B} = \{x_1, x_2, x_3\}}
         {\X_{G}^{\B'} \given \X_{D}^{\B'} = \{x_1, x_2, x_3\}} \\
    &= \underbrace{0.040 \times 1.721}_{\{a, c, e\}} +
       \underbrace{0.036 \times 1.721}_{\{b, c, e\}} +
       \underbrace{0.040 \times 2.504}_{\{a, d, e\}} +
       \underbrace{0.084 \times 2.504}_{\{b, d, e\}} + \\
    &\hspace{4em} \underbrace{0.16 \times 4.303}_{\{a, c, f\}} +
       \underbrace{0.144 \times 4.303}_{\{b, c, f\}} +
       \underbrace{0.16 \times 6.31}_{\{a, d, f\}} +
       \underbrace{0.336 \times 6.31}_{\{b, d, f\}} \\
    &= 4.879
  \end{align*}
  and $\KL{\B}{\B'} = \KL{\X_{D}^{\B}}{\X_{D}^{\B'}} +
  \KL{\X_{G}^{\B} \given \X_{D}^{\B}}{\X_{G}^{\B'} \given \X_{D}^{\B'}} =
  0.577 + 4.879 = 5.456$.

\label{ex:cgbn-kl}
\end{Example}

The computational complexity of this basic approach to computing $\KL{\B}{\B'}$
is
\begin{multline}
  \underbrace{
    \O{Mwl^{w + c} + M(w + wl^w + l^{w - 1}) + (Ml^c + 2)|\Theta_{\X_{D}}|}
  }_{\text{compute $\KL{\X_{D}^{\B}}{\X_{D}^{\B'}}$}} + \\
  \underbrace{
    \O{l^M \cdot \left(6(N - M)^3 + (N - M)^2 + 5(N - M)\right)}
  }_{\text{compute all the $\KL{\X_{G}^{\B} \given \X_{D}^{\B} =
       \x_{D}}{\X_{G}^{\B'} \given \X_{D}^{\B'} = \x_{D}}$}},
\label{eq:cgbn-kl-dense-big0}
\end{multline}
which we obtain by adapting \mref{eq:dbn-kl-big0} and \mref{eq:klg-big0} to
follow the notation $\lvert\X_{D}\rvert = M$ and $\lvert\X_{G}\rvert = N - M$ we
established in Section~\ref{sec:cgbn}. The first term implicitly covers the cost
of computing the $\Prob{\X_{D}^{\B} = \x_{D}}$, which relies on exact inference
like the computation of $\KL{\X_{D}^{\B}}{\X_{D}^{\B'}}$. The second term is
exponential in $M$, which would lead us to conclude that it is computationally
unfeasible to compute $\KL{\B}{\B'}$ whenever we have more than a few discrete
variables in $\B$ and $\B'$. Certainly, this would agree with \citet{gausmix},
who reviewed various scalable approximations of the KL divergence between two
Gaussian mixtures.

However, we would again disregard the fact that BNs are sparse models. Two
properties of CLGBNs that are apparent from Examples~\ref{ex:cgbn-compose}
and~\ref{ex:cgbn-kl} allow us to compute \mref{eq:cgbn-kl-2nd} efficiently:
\begin{itemize}
  \item We can reduce $\X_{G}^{\B} \given \X_{D}^{\B}$ to $\X_{G}^{\B} \given
    \Db{\B}$ where $\Db{\B} = \bigcup_{X_i \in \X_G} \DXi^{\B} \subseteq
    \X_{D}^{\B}$. In other words, the continuous nodes are conditionally
    independent on the discrete nodes that are not their parents ($\X_{D}^{\B}
    \setminus \Db{\B}$) given their parents ($\Db{\B}$). The same is true for
    $\X_{G}^{\B'} \given \X_{D}^{\B'}$. The number of distinct terms in the
    summation in \mref{eq:cgbn-kl-2nd} is then given by
    $\lvert \Val(\Db{\B} \cup \Db{\B'}) \rvert$ which will be smaller than
    $\lvert \Val(\X_{D}^{\B}) \rvert$ in sparse networks.
  \item The conditional distributions $\X_{G}^{\B} \given \X_{D}^{\B} = \db$ and
    $\X_{G}^{\B'} \given \X_{D}^{\B'} = \db$ are multivariate normals (not
    mixtures). They are also faithful to the subgraphs spanning the continuous
    nodes $\X_G$, and we can represent them as GBNs whose parameters can be
    extracted directly from $\B$ and $\B'$. Therefore, we can use the results
    from Section~\ref{sec:gbn-kl} to compute their Kullback-Leibler divergences
    efficiently.
\end{itemize}
As a result, \mref{eq:cgbn-kl-2nd} simplifies to
\begin{multline*}
\KL{\X_{G}^{\B} \given \X_{D}^{\B}}{\X_{G}^{\B'} \given \X_{D}^{\B'}} = \\
  \sum_{\db \in \Val(\Db{\B} \cup \Db{\B'})}
  \Prob{\{\Db{\B} \cup \Db{\B'}\} = \db}
  \KL{\X_{G}^{\B} \given \{\Db{\B} \cup \Db{\B'}\} = \db}
     {\X_{G}^{\B'} \given \{\Db{\B} \cup \Db{\B'}\} = \db}.
\end{multline*}
where $\Prob{\{\Db{\B} \cup \Db{\B'}\} = \db}$ is the probability that the nodes
$\Db{\B} \cup \Db{\B'}$ take value $\db$ as computed in $\B$. In turn,
\mref{eq:cgbn-kl-dense-big0} reduces to
\begin{multline*}
  \underbrace{
    \O{Mwl^{w + c} + M(w + wl^w + l^{w - 1}) + (Ml^c + 2)|\Theta_{\X_{D}}|}
  }_{\text{compute $\KL{\X_{D}^{\B}}{\X_{D}^{\B'}}$}} + \\
  \underbrace{
    \O{l^{|\Val(\Db{\B} \cup \Db{\B'})|} \cdot
      \left((N - M)^3 + 5(N - M)^2 + 6(N - M)\right)}
  }_{\text{compute all the
      $\KL{\X_{G}^{\B} \given \{\Db{\B} \cup \Db{\B'}\} = \db}
          {\X_{G}^{\B'} \given \{\Db{\B} \cup \Db{\B'}\} = \db}$}}.
\end{multline*} \label{page:cgbn-kl-sparse-big0}
because we can replace $l^M$ with $l^{|\Val(\Db{\B} \cup \Db{\B'})|}$, which is
an upper bound to the unique components in the mixture, and because we replace
the complexity in \mref{eq:klg-big0} with that \mref{eq:klg-fast0}. We can also
further reduce the second term to quadratic complexity as we discussed in
Section~\ref{sec:gbn-kl}. The remaining drivers of the computational complexity
are:
\begin{itemize}
  \item the maximum clique size $w$ in the subgraph spanning $\X_{D}^{\B}$;
  \item the number of arcs from discrete nodes to continuous nodes in both $\B$
    and $\B'$ and the overlap between $\Db{\B}$ and $\Db{\B'}$.
\end{itemize}

\begin{Example}[Sparse KL between two CLGBNs]
  Consider again the CLGBNs $\B$ and $\B'$ from Example~\ref{ex:cgbn-kl}. The
  node sets $\Db{\B} = \{X_2, X_3 \}$ and $\Db{\B'} = \{ X_3 \}$ identify
  four KL divergences to compute: $\Val(\Db{\B} \cup \Db{\B'}) =
  \left\{\{c, e\}, \{c, f\}, \{d, e\}, \{d, f\}\right\}$.
  \begin{multline*}
    \KL{\X_{G}^{\B} \given \X_{D}^{\B}}{\X_{G}^{\B'} \given \X_{D}^{\B'}} = \\
      \Prob{\{\Db{\B} \cup \Db{\B'}\} = \{c, e\}}
      \KL{\X_{G}^{\B} \given \{\Db{\B} \cup \Db{\B'}\} = \{c, e\}}
         {\X_{G}^{\B'} \given \{\Db{\B} \cup \Db{\B'}\} = \{c, e\}} + \\
      \Prob{\{\Db{\B} \cup \Db{\B'}\} = \{c, f\}}
      \KL{\X_{G}^{\B} \given \{\Db{\B} \cup \Db{\B'}\} = \{c, f\}}
         {\X_{G}^{\B'} \given \{\Db{\B} \cup \Db{\B'}\} = \{c, f\}} + \\
      \Prob{\{\Db{\B} \cup \Db{\B'}\} = \{d, e\}}
      \KL{\X_{G}^{\B} \given \{\Db{\B} \cup \Db{\B'}\} = \{d, e\}}
         {\X_{G}^{\B'} \given \{\Db{\B} \cup \Db{\B'}\} = \{d, e\}} + \\
      \Prob{\{\Db{\B} \cup \Db{\B'}\} = \{d, f\}}
      \KL{\X_{G}^{\B} \given \{\Db{\B} \cup \Db{\B'}\} = \{d, f\}}
         {\X_{G}^{\B'} \given \{\Db{\B} \cup \Db{\B'}\} = \{d, f\}}
  \end{multline*}
  All the BNs in the Kullback-Leibler divergences are GBNs whose structure and
  local distributions can be read from $\B$ and $\B'$. The four GBNs associated
  with $\X_{G}^{\B} \given \{\Db{\B} \cup \Db{\B'}\}$ have nodes $\X_{G}^{\B} =
  \{ X_4, X_5, X_6 \}$, arcs $\{X_5 \rarr X_4, X_4 \rarr X_6\}$ and the local
  distributions listed in Figure~\ref{fig:cgbns}. The corresponding GBNs
  associated with $\X_{G}^{\B'} \given \{\Db{\B} \cup \Db{\B'}\}$ are, in fact,
  only two distinct GBNs associated with $\{e\}$ and $\{f\}$. They have arcs
  $\{X_4 \rarr X_6, X_5 \rarr X_6 \}$ and local distributions: for $\{e\}$,
  \begin{align*}
    &X_4 = 0.3 + \eX{4},& &\eX{4} \sim N(0, 0.16), \\
    &X_5 = 1.4 + \eX{5},& &\eX{5} \sim N(0, 1.69), \\
    &X_6 = 0.1 + 0.2 X_4 + 0.7 X_5 + \eX{6},& &\eX{6} \sim N(0, 1.44);
  \end{align*}
  for $\{f\}$,
  \begin{align*}
    &X_4 = 1.0 + \eX{4},& &\eX{4} \sim N(0, 0.09), \\
    &X_5 = 0.5 + \eX{5},& &\eX{5} \sim N(0, 2.25), \\
    &X_6 = 0.1 + 0.2 X_4 + 0.7 X_5 + \eX{6},& &\eX{6} \sim N(0, 1.44).
  \end{align*}
  Plugging in the numbers,
  \begin{multline*}
    \KL{\X_{G}^{\B} \given \X_{D}^{\B}}{\X_{G}^{\B'} \given \X_{D}^{\B'}} =
      \underbrace{0.076 \times 1.721}_{\{c, e\}} +
      \underbrace{0.304 \times 4.303}_{\{c, f\}} + \\
      \underbrace{0.124 \times 2.504}_{\{d, e\}} +
      \underbrace{0.496 \times 6.310}_{\{d, f\}} = 4.879
  \end{multline*}
  which matches the value we computed in Example~\ref{ex:cgbn-kl}.

\label{ex:cgbn-kl-sparse}
\end{Example}

\section{Conclusions}
\label{sec:conclusions}

We started this paper by reviewing the three most common distributional
assumptions for BNs: discrete BNs, Gaussian BNs (GBNs) and conditional linear
Gaussian BNs (CLGBNs). Firstly, we reviewed the link between the respective
global and local distributions, and we formalised the computational complexity of
decomposing the former into the latter (and vice versa).

We then leveraged these results to study the complexity of computing Shannon's
entropy. We can, of course, compute the entropy of a BN from its global
distribution using standard results from the literature. (In the case of
discrete BNs and CLGBNS, only for small networks because $\lvert\Theta\rvert$
grows combinatorially.) However, this is not computationally efficient because
we incur the cost of composing the global distribution. While the entropy does
not decompose along with the local distributions for either discrete BNs or
CLGBNS, we show that it is nevertheless efficient to compute it from them.

Computing the Kullback-Leibler divergence between two BNs following the little
material found in the literature is more demanding. The discrete case has been
thoroughly investigated by \citet{moral}. However, the literature typically
relies on composing the global distributions for GBNs and CGBNs. Using the local
distributions, thus leveraging the intrinsic sparsity of BNs, we showed how to
compute the Kullback-Leibler divergence exactly with greater efficiency.
For GBNs, we showed how to compute the Kullback-Leibler divergence approximately
with quadratic complexity (instead of cubic). If the two GBNs have compatible
node orderings and their parameters are estimated from the same data, we can
also approximate their Kullback-Leibler divergence with complexity that scales
with the number of parents of each node. All these results are summarised in
Appendix~\ref{app:bigO}.

Finally, we provided step-by-step numeric examples of how to compute Shannon's
entropy and the Kullback-Leibler divergence for discrete BNs, GBNs and CLGBNs.
(See also Appendix~\ref{app:examples}.) Considering this is a highly technical
topic, and no such examples are available anywhere in the literature, we feel
that they are helpful in demystifying this topic and in integrating BNs into
many general machine learning approaches.

\funding{This research received no external funding.}

\conflictsofinterest{The authors declare no conflict of interest.}

\pagebreak

\appendixtitles{no}
\appendixstart
\appendix

\section[\appendixname~\thesection]{Computational Complexity Results}
\label{app:bigO}

For ease of reference, we summarise here all the computational complexity
results in this paper, including the type of BN and the page where they have
been derived.

\begin{center}
\renewcommand{\arraystretch}{1.75}
\begin{tabular}{lll}
  \hline
  \multicolumn{3}{c}{Composing and decomposing the global distributions} \\
  \hline
  $\O{Nl^N + l \sum_{i = 1}^N l^{|\PX{i}|}}$ &
  discrete BNs & page~\pageref{eq:marg-comp} \\
  $\O{N^3}$ &
  GBNs & page~\pageref{page:gbn-comp} \\
  $\O{Ml^M + (N - M)^3 l^{\Db{}}}$ &
  CLGBNs & page~\pageref{page:cgbn-comp} \\

  \hline
  \multicolumn{3}{c}{Computing Shannon's entropy} \\
  \hline
  $\O{N(w(1 + l^w) + l^{w - 1}) + \lvert\Theta\rvert}$ &
  discrete BNs & page~\pageref{page:dbn-h-big0} \\
  $\O{N}$ &
  GBNs & page~\pageref{page:gbn-h-big0} \\
  $\O{\sum_{X_i \in \X_{D}} \lvert \TXi \rvert +
      \sum_{X_i \in \X_{G}} \max\left\{1, \lvert\VDXi\rvert \right\}}$ &
  CLGBNs & page~\pageref{page:cgbn-h-big0} \\

  \hline
  \multicolumn{3}{c}{Computing the Kullback-Leibler divergence} \\
  \hline
  $\O{N^2 w l^{w + c} + N(w + wl^w + l^{w - 1}) +
    (N l^c + 2) \lvert\Theta\rvert}$ &
  discrete BNs & page~\pageref{eq:dbn-kl-big0} \\
  $\O{6N^3 + N^2 + 5N}$ &
  GBNs & page~\pageref{eq:klg-big0} \\
  $\O{Mwl^{w + c} + M(w + wl^w + l^{w - 1}) + (Ml^c + 2)|\Theta_{\X_{D}}|} +$
  & \\
  $\hspace{1em}\O{l^M \cdot \left(6(N - M)^3 + (N - M)^2 + 5(N - M)\right)}$ &
  CLGBNs & page~\pageref{eq:cgbn-kl-dense-big0} \\

  \hline
  \multicolumn{3}{c}{Sparse Kullback-Leibler divergence} \\
  \hline
  $\O{N^3 + 5N^2 + 6N}$ &
  GBNs & page~\pageref{eq:klg-fast0} \\
  $\O{Mwl^{w + c} + M(w + wl^w + l^{w - 1}) + (Ml^c + 2)|\Theta_{\X_{D}}|} +$ & \\
  $\hspace{1em}\O{l^{|\Val(\Db{\B} \cup \Db{\B'})|} \cdot
     \left((N - M)^3 + 5(N - M)^2 + 6(N - M)\right)}$ &
  CLGBNs & page~\pageref{page:cgbn-kl-sparse-big0} \\[0.25em]

  \hline
  \multicolumn{3}{c}{Approximate Kullback-Leibler divergence} \\
  \hline
  $\O{7N^2 + 6N}$ &
  GBNs & page~\pageref{eq:trace-ub} \\

  \hline
  \multicolumn{3}{c}{Efficient empirical Kullback-Leibler divergence} \\
  \hline
  $\O{nN(2c + \sfrac{5}{2})}$ &
  GBNs & page~\pageref{page:gbn-kl-lrm-big0} \\
  \hline
\end{tabular}
\end{center}

\renewcommand{\theexample}{B.\arabic{example}}
\section[\appendixname~\thesection]{Additional Examples}
\label{app:examples}

\begin{Example}[Composing and decomposing a discrete BN]
  Consider the discrete BN $\B$ shown in Figure~\ref{fig:dbns} (top). Composing
  its global distribution entails computing the joint probabilities of all
  possible states of all variables,
  \begin{equation*}
    \{a, b\} \times \{c, d\} \times \{e, f\} \times \{g, h\},
  \end{equation*}
  and arranging them in the following four-dimensional probability table in
  which each dimension is associated with one of the variables.
  \begin{center}
    \small
    \setlength\extrarowheight{2.5pt}
    \begin{tabular}{r|rrp{0em}r|rrp{0em}r|rrp{0em}r|rr}
      \multicolumn{7}{c}{$X_1 = a$} & & \multicolumn{7}{c}{$X_1 = b$} \\
      \cline{1-7} \cline{9-15}
      \multicolumn{3}{c}{$X_2 = c$} & & \multicolumn{3}{c}{$X_2 = d$} & &
      \multicolumn{3}{c}{$X_2 = c$} & & \multicolumn{3}{c}{$X_2 = d$} \\
      \cline{1-3} \cline{5-7} \cline{9-11} \cline{13-15}
      \multicolumn{1}{c}{} & \multicolumn{2}{c}{$X_3$} & & \multicolumn{1}{c}{} & \multicolumn{2}{c}{$X_3$} & &
      \multicolumn{1}{c}{} & \multicolumn{2}{c}{$X_3$} & & \multicolumn{1}{c}{} & \multicolumn{2}{c}{$X_3$} \\
      $X_4$ & $e$ & $f$ & & $X_4$ & $e$ & $f$ & & $X_4$ & $e$ & $f$ & & $X_4$ & $e$ & $f$ \\
      \cline{1-3} \cline{5-7} \cline{9-11} \cline{13-15}
      $g$ & $0.005$ & $0.064$ & & $g$ & $0.052$ & $0.037$ & & $g$ & $0.013$ & $0.040$ & & $g$ & $0.050$ & $0.026$ \\
      $h$ & $0.022$ & $0.089$ & & $h$ & $0.210$ & $0.051$ & & $h$ & $0.051$ & $0.056$ & & $h$ & $0.199$ & $0.036$ \\
    \end{tabular}
  \end{center}
  The joint probabilities are computed by multiplying the appropriate cells of
  the CPTs, for instance
  \begin{multline*}
    \Prob{\X = \{a, d, f, h\}} = \\
      \Prob{X_1 = a} \Prob{X_2 = d} \Prob{X_3 = f \given X_1 = a, X_2 = d}
      \Prob{X_4 = h \given X_3 = f} = \\
      0.53 \cdot 0.66 \cdot 0.25 \cdot 0.58 = 0.051.
  \end{multline*}

  Conversely, we can decompose the global distribution into the local
  distributions by summing over all variables other than the nodes and their
  parents. For $X_1$, this means
  \begin{align*}
    \Prob{X_1 = a}
      &= \sum_{x_2 \in \{c, d\}} \sum_{x_3 \in \{e, f\}} \sum_{x_4 \in \{g, h\}}
          \Prob{X_1 = a, X_2 = x_2, X_3 = x_3, X_4 = x_4} \\
      &= 0.005 + 0.064 + 0.022 + 0.089 + 0.052 + 0.037 + 0.210 + 0.051 = 0.53, \\
    \Prob{X_1 = b}
      &= \sum_{x_2 \in \{c, d\}} \sum_{x_3 \in \{e, f\}} \sum_{x_4 \in \{g, h\}}
          \Prob{X_1 = b, X_2 = x_2, X_3 = x_3, X_4 = x_4} \\
      &= 0.013 + 0.040 + 0.051 + 0.056 + 0.050 + 0.026 + 0.199 + 0.036 = 0.47.
  \end{align*}
  Similarly, for $X_2$ we get
  \begin{align*}
    \Prob{X_2 = c}
      &= \sum_{x_1 \in \{a, b\}} \sum_{x_3 \in \{e, f\}} \sum_{x_4 \in \{g, h\}}
          \Prob{X_1 = x_1, X_2 = c, X_3 = x_3, X_4 = x_4} \\
      &= 0.005 + 0.064 +0.022 + 0.089 + 0.013 + 0.040 + 0.051 + 0.056 = 0.34, \\
    \Prob{X_2 = d}
      &= \sum_{x_1 \in \{a, b\}} \sum_{x_3 \in \{e, f\}} \sum_{x_4 \in \{g, h\}}
          \Prob{X_1 = x_1, X_2 = d, X_3 = x_3, X_4 = x_4} \\
      &= 0.052 + 0.037 + 0.210 + 0.051 + 0.050 + 0.026 + 0.199 + 0.036 = 0.66.
  \end{align*}
  For $X_4$, we first compute the joint distribution of $X_4$ and $X_3$ by
  marginalising over $X_1$ and $X_2$,
  \begin{multline*}
    \underbrace{
      \bordermatrix{
        ~ & e & f \cr
        g & 0.005 & 0.064 \cr
        h & 0.022 & 0.089
      }
    }_{\{a, c\}} +
    \underbrace{
      \bordermatrix{
        ~ & e & f \cr
        g & 0.052 & 0.037 \cr
        h & 0.210 & 0.051
      }
    }_{\{a, d\}} +
    \underbrace{
      \bordermatrix{
        ~ & e & f \cr
        g & 0.013 & 0.040 \cr
        h &  0.051 & 0.056
      }
    }_{\{b, c\}} + \\
    \underbrace{
      \bordermatrix{
        ~ & e & f \cr
        g & 0.050 & 0.026 \cr
        h & 0.199 & 0.036
      }
    }_{\{b, d\}} =
    \bordermatrix{
      ~ & e & f \cr
      g & 0.120 & 0.167 \cr
      h & 0.481 & 0.232
    };
  \end{multline*}
  from which we obtain the CPT for $X_4 \given X_3$ by normalising its
  columns.

  As for $X_3$, we marginalise over $X_2$ to get the joint distribution of
  $X_3$, $X_1$ and $X_2$
  \begin{equation*}
    \resizebox{\linewidth}{!}{\bordermatrix{
      ~ & \{a, c\} & \{a, d\} & \{b, c\} & \{b, d\} \cr
      e & 0.005 + 0.022 = 0.027 & 0.052 + 0.210 = 0.262 &
          0.013 + 0.051 = 0.064 & 0.050 + 0.199 = 0.248 \cr
      f & 0.064 + 0.089 = 0.153 & 0.037 + 0.051 = 0.087 &
          0.040 + 0.056 = 0.096 & 0.026 + 0.036  = 0.062 \cr
    }}
  \end{equation*}
  and we obtain the CPT for $X_3 \given X_1, X_2$ by normalising its columns
  as we did earlier with $X_4$.

\label{ex:dbn}
\end{Example}

\begin{Example}[Composing and decomposing a CLGBN]
  Consider the CLGBN $\B$ from Figure~\ref{fig:cgbns} (top). The $M = 3$
  discrete variables at the top of the network have the joint distribution
  below:
  \begin{center}
  \begin{tabular}{cccccccc}
    \multicolumn{8}{c}{$\{X_1, X_2, X_3\}$} \\
    $\{a, c, e\}$ & $\{b, c, e\}$ & $\{a, d, e\}$ & $\{b, d, e\}$ &
    $\{a, c, f\}$ & $\{b, c, f\}$ & $\{a, d, f\}$ & $\{b, d, f\}$ \\
    \hline
    $0.040$ & $0.036$ & $0.040$ & $0.084$ &
    $0.160$ & $0.144$ & $0.160$ & $0.336$ \\
  \end{tabular}
  \end{center}
  Its elements identify the components of the mixture that make up the global
  distribution of $\B$, and the associated probabilities are the probabilities
  of those components.

  We can then identify which parts of the local distributions of the $N - M = 3$
  continuous variables ($X_4$, $X_5$ and $X_6$) we need to compute
  $\Prob{X_4, X_5, X_6 \given X_1, X_2, X_3}$ for each element of the mixture.
  The graphical structure of $\B$ implies that
  \mbox{$\Prob{X_4, X_5, X_6 \given X_1, X_2, X_3} =$}
  \mbox{$\Prob{X_4, X_5, X_6 \given X_2, X_3}$} because the continuous nodes are
  \mbox{d-separated} from $X_1$ by their parents. As a result, the following
  mixture components will share identical distributions which only depend on the
  configurations of $X_2$ and $X_3$:
  \begin{align*}
    &\{a, c, e\}, \{b, c, e\} \mapsto \{c, e\},&
    &\{a, d, e\}, \{b, d, e\} \mapsto \{d, e\}, \\
    &\{a, c, f\}, \{b, c, f\} \mapsto \{c, f\},&
    &\{a, d, f\}, \{b, d, f\} \mapsto \{d, f\}.
  \end{align*}

  For the mixture components with a distribution identified by $\{c, e\}$, the
  relevant parts of the distributions of $X_4$, $X_5$ and $X_6$ are:
  \begin{align*}
    &X_4 = 0.1 + 0.2 X_5 + \eX{4},& &\eX{4} \sim N(0, 0.09); \\
    &X_5 = 0.1 + \eX{5},& &\eX{5} \sim N(0, 0.09); \\
    &X_6 = 0.1 + 0.2 X_4 + \eX{6},& &\eX{6} \sim N(0, 1).
  \end{align*}
  We can treat them as the local distributions in a GBN over $\{X_4, X_5,
  X_6\}$ with a DAG equal to the subgraph of $\B$ spanning only these nodes. If
  we follow the steps outlined in Section~\ref{sec:gbn} and illustrated in
  Example~\ref{ex:gbn}, we get
  \begin{align*}
    \begin{bmatrix} X_4 \\ X_5 \\ X_6 \end{bmatrix} \sim
      N\left(
        \begin{bmatrix} 0.120 \\ 0.100 \\ 0.124 \end{bmatrix},
        \SmB{\{c, e\}} = \begin{bmatrix}
          0.094 & 0.018 & 0.019 \\
          0.018 & 0.090 & 0.004 \\
          0.019 & 0.004 & 1.004
        \end{bmatrix}
      \right)
  \end{align*}
  which is the multivariate normal distribution associated with the components
  $\{a, c, e\}$ and $\{b, c, e\}$ in the mixture. Similarly, the relevant parts
  of the distributions of $X_4$, $X_5$ and $X_6$ for $\{d, e\}$ are
  \begin{align*}
    &X_4 = 0.6 + 0.8 X_5 + \eX{4},& &\eX{4} \sim N(0, 0.36); \\
    &X_5 = 0.2 + \eX{5},& &\eX{5} \sim N(0, 0.36); \\
    &X_6 = 0.1 + 0.2 X_4 + \eX{6},& &\eX{6} \sim N(0, 1);
  \end{align*}
  and jointly
  \begin{align*}
    \begin{bmatrix} X_4 \\ X_5 \\ X_6 \end{bmatrix} \sim
      N\left(
        \begin{bmatrix} 0.760 \\ 0.200 \\ 0.252 \end{bmatrix},
        \SmB{\{d, e\}} = \begin{bmatrix}
          0.590 & 0.288 & 0.118 \\
          0.288 & 0.360 & 0.058 \\
          0.118 & 0.058 & 1.024
        \end{bmatrix}
      \right)
  \end{align*}
  for the components $\{a, d, e\}$ and $\{b, d, e\}$. For the components
  $\{a, c, f\}$ and $\{b, c, f\}$, the local distributions identified by
  $\{c, f\}$ are
  \begin{align*}
    &X_4 = 0.1 + 0.2 X_5 + \eX{4},& &\eX{4} \sim N(0, 0.09); \\
    &X_5 = 0.4 + \eX{5},& &\eX{5} \sim N(0, 0.81); \\
    &X_6 = 0.1 + 0.2 X_4 + \eX{6},& &\eX{6} \sim N(0, 1);
  \end{align*}
  and the joint distribution of $X_4$, $X_5$ and $X_6$ is
  \begin{align*}
    \begin{bmatrix} X_4 \\ X_5 \\ X_6 \end{bmatrix} \sim
      N\left(
        \begin{bmatrix} 0.180 \\ 0.400 \\ 0.136 \end{bmatrix},
        \SmB{\{c, f\}} = \begin{bmatrix}
          0.122 & 0.162 & 0.024 \\
          0.162 & 0.810 & 0.032 \\
          0.024 & 0.032 & 1.005
        \end{bmatrix}
      \right).
  \end{align*}
  Finally, the local distributions identified by $\{d, f\}$ are
  \begin{align*}
    &X_4 = 0.6 + 0.8 X_5 + \eX{4},& &\eX{4} \sim N(0, 0.36); \\
    &X_5 = 0.4 + \eX{5},& &\eX{5} \sim N(0, 1.44); \\
    &X_6 = 0.1 + 0.2 X_4 + \eX{6},& &\eX{6} \sim N(0, 1);
  \end{align*}
  and the joint distribution of $X_4$, $X_5$ and $X_6$ for the components
  $\{a, d, f\}$, $\{b, d, f\}$ is
  \begin{align*}
    \begin{bmatrix} X_4 \\ X_5 \\ X_6 \end{bmatrix} \sim
      N\left(
        \begin{bmatrix} 0.920 \\ 0.400 \\ 0.284 \end{bmatrix},
        \SmB{\{d, f\}} = \begin{bmatrix}
          1.282 & 1.152 & 0.256 \\
          1.152 & 1.440 & 0.230 \\
          0.256 & 0.230 & 1.051
        \end{bmatrix}
      \right).
  \end{align*}

  We follow the same steps in reverse to decompose the global distribution into
  the local distributions. The joint distribution of $\X$ is a mixture with
  multivariate normal components and the associated probabilities. The latter
  are a function of the discrete variables $X_1$, $X_2$, $X_3$: rearranging them
  as the three-dimensional table
  \begin{center}
  \begin{tabular}{ll|cc}
    \multicolumn{4}{c}{$X_1 = a$} \\
    \toprule
    \multicolumn{2}{c}{} & \multicolumn{2}{c}{$X_2$} \\
    & & $c$ & $d$ \\
    \cline{2-4}
    \multirow{2}{*}{$X_3$}
    & $e$ & $0.040$ & $0.040$ \\
    & $f$ & $0.160$ & $0.160$ \\
  \end{tabular}
  \hspace{0.10\linewidth}
  \begin{tabular}{ll|cc}
    \multicolumn{4}{c}{$X_1 = b$} \\
    \toprule
    \multicolumn{2}{c}{} & \multicolumn{2}{c}{$X_2$} \\
    & & $c$ & $d$ \\
    \cline{2-4}
    \multirow{2}{*}{$X_3$}
    & $e$ & $0.036$ & $0.084$ \\
    & $f$ & $0.144$ & $0.336$ \\
  \end{tabular}
  \end{center}
  gives us the typical representation of $\Prob{X_1, X_2, X_3}$, which we can
  work with by operating over the different dimensions. We can then compute
  the conditional probability tables in the local distributions of $X_1$ and
  $X_3$ by marginalising over the remaining variables:
  \begin{align*}
    \Prob{X_1}
      &= \sum_{X_2 \in \{c,d\}} \sum_{X_3 \in \{e, f\}} \Prob{X_1, X_2, X_3} \\
      &= \bordermatrix{
        ~ & a & b \cr
        ~ & 0.040 + 0.160 + 0.040 + 0.160 &
            0.036 + 0.144 + 0.084 + 0.336
      } \\
      &= \bordermatrix{
        ~ & a & b \cr
        ~ & 0.4 & 0.6
      }, \\
    \Prob{X_3}
      &= \sum_{X_1 \in \{a, b\}} \sum_{X_2 \in \{c,d\}} \Prob{X_1, X_2, X_3} \\
      &= \bordermatrix{
        ~ & e & f \cr
        ~ & 0.040 + 0.040 + 0.036 + 0.084 &
            0.160 + 0.160 + 0.144 + 0.336
      } \\
      &= \bordermatrix{
        ~ & e & f \cr
        ~ & 0.2 & 0.8
      }.
  \end{align*}
  As for $X_2$, we marginalise over $X_3$ and normalise over $X_1$ to obtain
  \begin{equation*}
    \Prob{X_2 \given X_1}
      = \sum_{X_3 \in \{e, f\}} \frac{\Prob{X_1, X_2, X_3}}{\Prob{X_1}}
      = \bordermatrix{
        ~ & a & b \cr
        c & \frac{0.040 + 0.160}{0.4} & \frac{0.036 + 0.144}{0.6} \cr
        d & \frac{0.040 + 0.160}{0.4} & \frac{0.084 + 0.336}{0.6}
      }
      = \bordermatrix{
        ~ & a & b \cr
        c & 0.5 & 0.3 \cr
        d & 0.5 & 0.7
      }.
  \end{equation*}
  The multivariate normal distributions associated with the mixture components
  are a function of the continuous variables $X_4$, $X_5$, $X_6$. $X_4$ has only
  one discrete parent ($X_2$), $X_5$ has two ($X_2$ and $X_3$) and $X_6$ has
  none. Therefore, we only need to examine four mixture components to obtain the
  parameters of the local distributions of all three variables: one for which
  $\{X_2 = c, X_3 = e\}$, one for which $\{X_2 = d, X_3 = e\}$, one for which
  $\{X_2 = c, X_3 = f\}$ and one for which $\{X_2 = d, X_3 = f\}$.

  If we consider the first mixture component $\{a, c, e\}$, we can apply the
  steps described Section~\ref{sec:gbn} to decompose it into the local
  distributions of $X_4$, $X_5$, $X_6$ and get
  \begin{align*}
    &X_4 = 0.1 + 0.2 X_5 + \eX{4},& &\eX{4} \sim N(0, 0.09); \\
    &X_5 = 0.1 + \eX{5},          & &\eX{5} \sim N(0, 0.09); \\
    &X_6 = 0.1 + 0.2 X_4 + \eX{6},& &\eX{6} \sim N(0, 1).
  \end{align*}
  Similarly, the third mixture component $\{a, d, e\}$ yields
  \begin{align*}
    &X_4 = 0.6 + 0.8 X_5 + \eX{4},& &\eX{4} \sim N(0, 0.36); \\
    &X_5 = 0.2 + \eX{5},          & &\eX{5} \sim N(0, 0.36); \\
    &X_6 = 0.1 + 0.2 X_4 + \eX{6},& &\eX{6} \sim N(0, 1).
  \end{align*}
  The fifth mixture component $\{a, c, f\}$ yields
  \begin{align*}
    &X_4 = 0.1 + 0.2 X_5 + \eX{4},& &\eX{4} \sim N(0, 0.09); \\
    &X_5 = 0.4 + \eX{5},          & &\eX{5} \sim N(0, 0.81); \\
    &X_6 = 0.1 + 0.2 X_4 + \eX{6},& &\eX{6} \sim N(0, 1).
  \end{align*}
  The seventh mixture component $\{a, d, f\}$ yields
  \begin{align*}
    &X_4 = 0.6 + 0.8 X_5 + \eX{4},& &\eX{4} \sim N(0, 0.36); \\
    &X_5 = 0.4 + \eX{5},          & &\eX{5} \sim N(0, 1.44); \\
    &X_6 = 0.1 + 0.2 X_4 + \eX{6},& &\eX{6} \sim N(0, 1).
  \end{align*}
  Reorganising these distributions by variables we obtain the local
  distributions of $\B$ shown in Figure~\ref{fig:cgbns} (top).

\label{ex:cgbn}
\end{Example}

\begin{Example}[Entropy of a discrete BN]
  Consider again the discrete BN from Example~\ref{ex:dbn}. In this simple
  example, we can use its global distribution and \mref{eq:entropy} to compute
  \begin{multline*}
    \HH{\B} =
      - 0.005 \log 0.005 - 0.013 \log 0.013 - 0.052 \log 0.052 -
        0.050 \log 0.050 - \\ 0.064 \log 0.064 - 0.040 \log 0.040 -
        0.037 \log 0.037 - 0.026 \log 0.026 - \\ 0.022 \log 0.022 -
        0.051 \log 0.051 - 0.210 \log 0.210 - 0.199 \log 0.199 - \\
        0.089 \log 0.089 - 0.056 \log 0.056 - 0.051 \log 0.051 -
        0.036 \log 0.036 = 2.440.
  \end{multline*}
  In the general case, we compute $\HH{\B}$ from the local distributions using
  \mref{eq:dbn-h}. Since $X_1$ and $X_2$ have no parents, their entropy
  components simply sum over their marginal distributions:
  \begin{align*}
    \HH{X_1} &= - 0.53 \log 0.53 - 0.47 \log 0.47 = 0.691, \\
    \HH{X_2} &= - 0.34 \log 0.34 - 0.66 \log 0.66 = 0.641.
  \end{align*}
  For $X_3$,
  \begin{equation*}
    \HH{X_3 \given X_1, X_2} =
      \sum_{x_1 \in \{a, b\}} \sum_{x_2 \in \{c, d\}}
        \Prob{X_1 = x_1, X_2 = x_2} \HH{X_3 \given X_1 = x_1, X_2 = x_2}
  \end{equation*}
  where
  \begin{align*}
    \HH{X_3 \given X_1 = a, X_2 = c}
      &= - 0.15 \log 0.15 - 0.85 \log 0.85 = 0.423, \\
    \HH{X_3 \given X_1 = a, X_2 = d}
      &= - 0.75 \log 0.75 - 0.25 \log 0.25 = 0.562, \\
    \HH{X_3 \given X_1 = b, X_2 = c}
      &= - 0.40 \log 0.40 - 0.60 \log 0.60 = 0.673, \\
    \HH{X_3 \given X_1 = b, X_2 = d}
      &= - 0.80 \log 0.80 - 0.20 \log 0.20 = 0.500;
  \end{align*}
  and where (multiplying the marginal probabilities for $X_1$ and $X_2$, which
  are marginally independent)
  \begin{align*}
    &\Prob{X_1 = a, X_2 = c} = 0.180, &
    &\Prob{X_1 = a, X_2 = d} = 0.350, \\
    &\Prob{X_1 = b, X_2 = c} = 0.160, &
    &\Prob{X_1 = b, X_2 = d} = 0.310;
  \end{align*}
  giving
  \begin{equation*}
    \HH{X_3 \given X_1, X_2} =
      (0.180 \cdot 0.423 + 0.350 \cdot 0.562 + 0.160 \cdot 0.673 +
       0.310 \cdot 0.500) = 0.536.
  \end{equation*}
  Finally, for $X_4$
  \begin{equation*}
    \HH{X_4 \given X_3} =
      \sum_{x_3 \in \{e, f\}} \Prob{X_3 = x_3} \HH{X_4 \given X_3 = x_3}
  \end{equation*}
  where
  \begin{align*}
    \HH{X_4 \given X_3 = e} &= - 0.20 \log 0.20 - 0.80 \log 0.80 = 0.500, \\
    \HH{X_4 \given X_3 = f} &= - 0.42 \log 0.42 - 0.58 \log 0.58 = 0.680;
  \end{align*}
  and $\Prob{X_3 = e} = 0.601$, $\Prob{X_3 = f} = 0.399$, giving
  \begin{equation*}
    \HH{X_4 \given X_3} = 0.601 \cdot 0.500 + 0.399 \cdot 0.680 = 0.572.
  \end{equation*}
  Combining all these figures, we obtain $\HH{\B}$ as
  \begin{equation*}
    \HH{X_1} + \HH{X_2} + \HH{X_3 \given X_1, X_2} + \HH{X_4 \given X_3} =
      0.691 + 0.641 + 0.536 + 0.572 = 2.440
  \end{equation*}
  as before.

  In general, we would have to compute the probabilities of the parent
  configurations of each node using a junction tree as follows:
  \begin{enumerate}
    \item We construct the moral graph of $\B$, which contains the same arcs
      (but undirected) as its DAG plus $X_1 \uarr X_2$.
    \item We identify two cliques $C_1 = \{X_1, X_2, X_3\}$ and
      $C_2 = \{X_3, X_4\}$ and a separator $S_{12} = \{X_3\}$.
    \item We connect them to create the junction tree
      $C_1 \uarr S_{12} \uarr C_2$.
    \item We initialise the cliques with the respective distributions
      $\Prob{C_1} = \Prob{X_1, X_2, X_3}$, \linebreak
      $\Prob{C_2} = \Prob{X_3, X_4}$ and $\Prob{S_{12}} = \Prob{X_3}$.
    \item We compute $\Prob{X_1, X_2} = \sum_{x_3 \in \{e, f\}} \Prob{C_1}$
      and $\Prob{X_3} = \Prob{S_{12}}$.
  \end{enumerate}

\label{ex:dbn-entropy}
\end{Example}

\begin{Example}[Entropy of a GBN]
  Consider the GBN $\B$ from Figure~\ref{fig:gbns} (top), whose global
  distribution we derived in Example~\ref{ex:gbn}. If we plug its covariance
  matrix $\SB$ into the entropy formula for the multivariate  normal
  distribution we obtain
  \begin{equation*}
    \HH{\B}
      = \frac{4}{2} + \frac{4}{2} \log 2\pi + \frac{1}{2} \log\det(\SB)
      = 2 + 3.676 + 0.5 \log 0.475 = 5.304.
  \end{equation*}
  Equivalently, plugging the $\sigma^2_{X_i}(\B)$ into \mref{eq:gbn-h} we have
  \begin{multline*}
    \HH{\B}
      = \sum_{i = 1}^N \HH{\XP{i}^{\B}} = \\
      \frac{1}{2} \left[
        \log(2\pi \cdot 0.8) + \log(2\pi \cdot 0.6) +
        \log(2\pi \cdot 0.9) + \log(2\pi \cdot 1.1)
      \right] + \frac{4}{2}
      = 5.304.
  \end{multline*}
\label{ex:gbn-entropy}
\end{Example}

\begin{figure}[t]
  \centering
  \includegraphics[height=0.35\textheight]{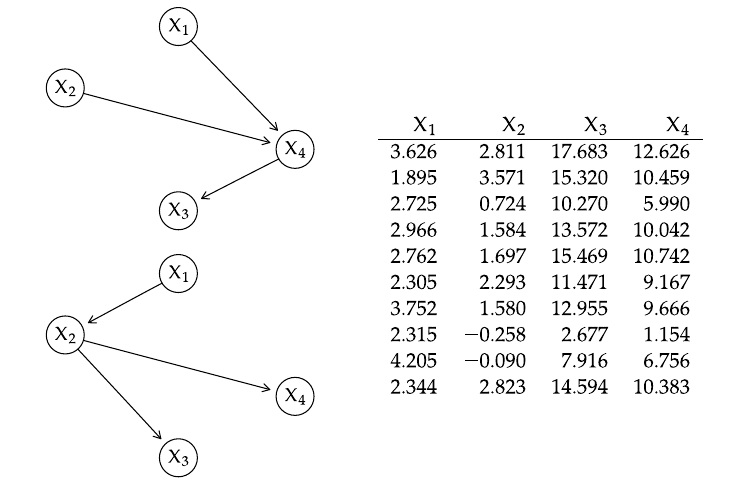}
  \caption{The DAGs for the GBNs $\B$ (top left) and $\B'$
    (bottom left) and the data (right) used in Example~\ref{ex:gbn-lrm}.}
  \label{fig:gbns-from-data}
\end{figure}

\begin{Example}[KL between GBNs with parameters estimated from data]
  Consider the DAGs for the BNs $\B$ and $\B'$ and the 10 observations shown in
  Figure~\ref{fig:gbns-from-data}. The partial topological ordering of the nodes
  in $\B$ is $\{\{X_1, X_2\}, X_4, X_3\}$ and that in $\B'$ is $\{X_1, X_2,
  \{X_3, X_4\}\}$: the total ordering that is compatible with both is $\{X_1,
  X_2, X_4, X_3\}$.

  If we estimate the parameters of the local distributions of $\B$ by maximum
  likelihood we get
  \begin{align*}
    &X_1 = 2.889 + \eX{1},& &\eX{1} \sim N(0, 0.558), \\
    &X_2 = 1.673 + \eX{2},& &\eX{2} \sim N(0, 1.595), \\
    &X_3 = 0.896 + 1.299 X_4 + \eX{3},& &\eX{3} \sim N(0, 1.142), \\
    &X_4 = -2.095 + 2.222 X_1 + 2.613 X_2 + \eX{4},& &\eX{4} \sim N(0, 1.523),
  \end{align*}
  and the associated fitted values are
  \begin{align*}
    \wx{1}(\B) &=
      (2.889, 2.889, 2.889, 2.889, 2.889, 2.889, 2.889, 2.889, 2.889, 2.889), \\
    \wx{2}(\B) &=
      (1.673, 1.673, 1.673, 1.673, 1.673, 1.673, 1.673, 1.673, 1.673, 1.673), \\
    \wx{3}(\B) &=
      (17.293, 14.480, 8.675, 13.937, 14.846, 12.801, 13.449, 2.394, 9.670, 14.381), \\
    \wx{4}(\B) &=
      (13.307, 11.447, 5.852, 8.635, 8.475, 9.018, 10.370, 2.376, 7.014, 10.489).
  \end{align*}
  Similarly, for $\B'$ we get
  \begin{align*}
    &X_1 = 2.889 + \eX{1},& &\eX{1} \sim N(0, 0.558), \\
    &X_2 = 3.505 - 0.634 X_1 + \eX{2},& &\eX{2} \sim N(0, 1.542), \\
    &X_3 = 7.284 + 2.933 X_2 + \eX{3},& &\eX{3} \sim N(0, 6.051), \\
    &X_4 = 5.151 + 2.120 X_2 + \eX{4},& &\eX{4} \sim N(0, 3.999),
  \end{align*}
  and the associated fitted values are
  \begin{align*}
    \wx{1}(\B) &=
      (2.889, 2.889, 2.889, 2.889, 2.889, 2.889, 2.889, 2.889, 2.889, 2.889), \\
    \wx{2}(\B) &=
      (1.207, 2.304, 1.778, 1.625, 1.754, 2.044, 1.127, 2.037, 0.840, 2.019), \\
    \wx{3}(\B) &=
      (15.529, 17.760, 9.408, 11.931, 12.261, 14.009, 11.918, 6.528, 7.019, 15.564), \\
    \wx{4}(\B) &=
      (11.110, 12.722, 6.686, 8.509, 8.748, 10.011, 8.500, 4.604, 4.959, 11.135).
  \end{align*}
  Therefore,
  \begin{align*}
    &\| \wx{1}(\B) - \wx{1}(\B') \|^2_2 = 0,&
    &\| \wx{2}(\B) - \wx{2}(\B') \|^2_2 = 2.018, \\
    &\| \wx{3}(\B) - \wx{3}(\B') \|^2_2 = 54.434,&
    &\| \wx{4}(\B) - \wx{4}(\B') \|^2_2 = 21.329;
  \end{align*}
  and the values of the Kullback-Leibler divergence for the individual nodes are
  \begin{align*}
    \KL{\XP{1}^{\B}}{\XP{1}^{\B'}} &\approx
      \frac{1}{2} \left( \log\frac{0.558}{0.558} + \frac{0.558}{0.558} - 1\right)
        + \frac{1}{20} \left( \frac{0}{0.558} \right) = 0, \\
    \KL{\XP{2}^{\B}}{\XP{2}^{\B'}} &\approx
      \frac{1}{2} \left( \log\frac{1.542}{1.595} + \frac{1.595}{1.542} - 1\right)
        + \frac{1}{20} \left( \frac{2.018}{1.542} \right) = 0.066, \\
    \KL{\XP{3}^{\B}}{\XP{3}^{\B'}} &\approx
      \frac{1}{2} \left( \log\frac{6.051}{1.142} + \frac{1.142}{6.051} - 1\right)
        + \frac{1}{20} \left( \frac{54.434}{6.051} \right) = 0.878, \\
    \KL{\XP{4}^{\B}}{\XP{4}^{\B'}} &\approx
      \frac{1}{2} \left( \log\frac{3.999}{1.523} + \frac{1.523}{3.999} - 1\right)
        + \frac{1}{20} \left( \frac{21.329}{3.999} \right) = 0.440,
  \end{align*}
  which sum up to $\KL{\B}{\B'} \approx 1.383$. The exact value, which we can
  compute as shown in Section~\ref{sec:gbn-kl}, is $1.692$.

  The quality of the empirical approximation improves with the number of
  observations. For reference, we generated the data in
  Figure~\ref{fig:gbns-from-data} from the GBN in Example~\ref{ex:gbn}. With a
  sample of size $n = 100$ from the same network, $\KL{\B}{\B'} \approx 1.362$
  with $\KL{\B}{\B'} = 1.373$; with $n = 1000$, $\KL{\B}{\B'} \approx 1.343$
  with $\KL{\B}{\B'} = 1.345$.

  \label{ex:gbn-lrm}
\end{Example}

\begin{Example}[Entropy of a CLGBN]
  Consider again the CLGBN $\B$ from from Figure~\ref{fig:cgbns} (top). For such
  a simple BN, we can use its global distribution (which we derived in
  Example~\ref{ex:cgbn}) directly to compute the entropies of the multivariate
  normal distributions associated with the mixture components
  \begin{align*}
    \HH{X_4, X_5, X_6 \given \{c, e\}} & =
      \frac{3}{2} + \frac{3}{2} \log(2\pi) +
      \frac{1}{2} \log\det\left(\SmB{\{c, e\}}\right) = 1.849, \\
    \HH{X_4, X_5, X_6 \given \{d, e\}} & =
      \frac{3}{2} + \frac{3}{2} \log(2\pi) +
      \frac{1}{2} \log\det\left(\SmB{\{d, e\}}\right) = 3.235, \\
    \HH{X_4, X_5, X_6 \given \{c, f\}} & =
      \frac{3}{2} + \frac{3}{2} \log(2\pi) +
      \frac{1}{2} \log\det\left(\SmB{\{c, f\}}\right) = 2.947, \\
    \HH{X_4, X_5, X_6 \given \{d, f\}} & =
      \frac{3}{2} + \frac{3}{2} \log(2\pi) +
      \frac{1}{2} \log\det\left(\SmB{\{d, f\}}\right) = 3.928;
  \end{align*}
  and to combine them by weighting with the component probabilities
  \begin{multline*}
    \HH{X_4, X_5, X_6 \given X_1, X_2, X_3} =
      \underbrace{0.040 \cdot 1.849}_{\{a, c, e\}} +
      \underbrace{0.036 \cdot 1.849}_{\{b, c, e\}} +
      \underbrace{0.040 \cdot 3.235}_{\{a, d, e\}} +
      \underbrace{0.084 \cdot 3.235}_{\{b, d, e\}} + \\
      \underbrace{0.160 \cdot 2.947}_{\{a, c, f\}} +
      \underbrace{0.144 \cdot 2.947}_{\{b, c, f\}} +
      \underbrace{0.160 \cdot 3.928}_{\{a, d, f\}} +
      \underbrace{0.336 \cdot 3.928}_{\{b, d, f\}} = 3.386.
  \end{multline*}
  The entropy of the discrete variables is
  \begin{multline*}
    \HH{X_1, X_2, X_3} =
    - 0.040 \log 0.040 - 0.036 \log 0.036 - 0.040 \log 0.040 -
      0.084 \log 0.084 - \\ 0.160 \log 0.160 - 0.144 \log 0.144 -
      0.160 \log 0.160 - 0.336 \log 0.336 = 1.817
  \end{multline*}
  and then $\HH{\B} = \HH{X_1, X_2, X_3} +
  \HH{X_4, X_5, X_6 \given X_1, X_2, X_3} = 5.203$.

  If we use the local distributions instead, we can compute the entropy of the
  discrete variables using \mref{eq:dbn-h} from Section~\ref{sec:dbn-kl}:
  \begin{align*}
    \HH{X_1} &= - 0.4 \log 0.4 - 0.6 \log 0.6 = 0.673, \\
    \HH{X_2 \given X_1} &= 0.4 (- 0.5 \log 0.5 - 0.5 \log 0.5) +
                           0.6 (- 0.3 \log 0.3 - 0.7 \log 0.7) = 0.644, \\
    \HH{X_3} &= - 0.2 \log 0.2 - 0.8 \log 0.8 = 0.500.
  \end{align*}
  We can compute the entropy of the continuous variables with no discrete parents
  using \mref{eq:gbn-h} from Section~\ref{sec:gbn-kl}:
  \begin{equation*}
    \HH{X_6 \given X_4} = \frac{1}{2} \log(2 \pi \cdot 1) + \frac{1}{2} = 1.419.
  \end{equation*}
  Finally, we can compute the entropy of the continuous variables with discrete
  parents using \mref{eq:cgbn-h} from Section~\ref{sec:cgbn-kl}:
  \begin{align*}
    \HH{X_4 \given X_2, X_5} &=
      0.38 \left(\frac{1}{2} \log(2 \pi \cdot 0.09) + \frac{1}{2}\right)+
      0.62 \left(\frac{1}{2} \log(2 \pi \cdot 0.36) + \frac{1}{2}\right) \\
      &= 0.645, \\
    \HH{X_5 \given X_2, X_3} &=
      0.076 \left(\frac{1}{2} \log(2 \pi \cdot 0.09) + \frac{1}{2}\right) +
      0.124 \left(\frac{1}{2} \log(2 \pi \cdot 0.36) + \frac{1}{2}\right) + \\
      &\phantom{.=}
      0.304 \left(\frac{1}{2} \log(2 \pi \cdot 0.81) + \frac{1}{2}\right) +
      0.496 \left(\frac{1}{2} \log(2 \pi \cdot 1.44) + \frac{1}{2}\right) \\
      &= 1.322.
  \end{align*}
  As before, we confirm that overall
  \begin{multline*}
    \HH{\B} = \HH{X_1} + \HH{X_2 \given X_1} + \HH{X_3} +
      \HH{X_4 \given X_2, X_5} + \HH{X_5 \given X_2, X_3} + \\
      \HH{X_6 \given X_4} = 0.673 + 0.644 + 0.500 + 0.645 + 1.322 + 1.419
      = 5.203.
  \end{multline*}

\label{ex:cgbn-entropy}
\end{Example}

\begin{adjustwidth}{-\extralength}{0cm}
\reftitle{References}


\begin{thebibliography}{999}

\bibitem[Scutari and Denis(2021)]{crc21}
Scutari, M.; Denis, J.B.
\newblock {\em {Bayesian Networks with Examples in R}}, 2nd ed.; Chapman \&
  Hall,  2021.

\bibitem[Castillo et~al.(1997)Castillo, Guti{\'e}rrez, and Hadi]{castillo}
Castillo, E.; Guti{\'e}rrez, J.M.; Hadi, A.S.
\newblock {\em {Expert Systems and Probabilistic Network Models}}; Springer,
  1997.

\bibitem[Cowell et~al.(1999)Cowell, Dawid, Lauritzen, and
  Spiegelhalter]{cowell2}
Cowell, R.G.; Dawid, A.P.; Lauritzen, S.L.; Spiegelhalter, D.J.
\newblock {\em {Probabilistic Networks and Expert Systems}}; Springer,  1999.

\bibitem[Pearl(1988)]{pearl}
Pearl, J.
\newblock {\em {Probabilistic Reasoning in Intelligent Systems: Networks of
  Plausible Inference}}; Morgan Kaufmann,  1988.

\bibitem[Koller and Friedman(2009)]{koller}
Koller, D.; Friedman, N.
\newblock {\em {Probabilistic Graphical Models: Principles and Techniques}};
  MIT Press,  2009.

\bibitem[Murphy(2002)]{murphy}
Murphy, K.P.
\newblock {Dynamic Bayesian Networks: Representation, Inference and Learning}.
\newblock PhD thesis, Computer Science Division, UC Berkeley,  2002.

\bibitem[Spirtes et~al.(2000)Spirtes, Glymour, and Scheines]{spirtes}
Spirtes, P.; Glymour, C.; Scheines, R.
\newblock {\em {Causation, Prediction, and Search}}; MIT Press,  2000.

\bibitem[Pearl(2009)]{causality}
Pearl, J.
\newblock {\em {Causality: Models, Reasoning and Inference}}, 2nd ed.;
  Cambridge University Press,  2009.

\bibitem[Borsboom et~al.(2021)Borsboom, Deserno, Rhemtulla, Epskamp, Fried,
  McNally, Robinaugh, Perugini, Dalege, Costantini, Isvoranu, Wysocki, {van
  Borkulo}, {van Bork}, and Waldorp]{mcnally}
Borsboom, D.; Deserno, M.K.; Rhemtulla, M.; Epskamp, S.; Fried, E.I.; McNally,
  R.J.; Robinaugh, D.J.; Perugini, M.; Dalege, J.; Costantini, G.;  et~al.
\newblock {Network Analysis of Multivariate Data in Psychological Science}.
\newblock {\em Nature Reviews Methods Primers} {\bf 2021}, {\em 1},~58.

\bibitem[Carapito et~al.(2021)Carapito, Li, Helms, Carapito, Gujja, Rolli,
  Guimaraes, Malagon-Lopez, Spinnhirny, Lederle, Mohseninia, Hirschler, Muller,
  Bastard, Gervais, Zhang, Danion, Ruch, Schenck, Collange, Chamaraux-Tran,
  Molitor, Pichot, Bernard, Tahar, Bibi-Triki, Wu, Paul, Mayeur, Larnicol,
  Laumond, Frappier, Schmidt, Hanauer, Macquin, Stemmelen, Simons, Mariette,
  Hermine, Fafi-Kremer, Goichot, Drenou, Kuteifan, Pottecher, Mertes, Kailasan,
  {Javad Aman}, Pin, Nilsson, Thomas, Viari, Sanlaville, Schneider, Sibilia,
  Tharaux, Casanova, Hansmann, Lidar, Radosavljevic, Gulcher, Meziani, Moog,
  Chittenden, and Bahram]{covid}
Carapito, R.; Li, R.; Helms, J.; Carapito, C.; Gujja, S.; Rolli, V.; Guimaraes,
  R.; Malagon-Lopez, J.; Spinnhirny, P.; Lederle, A.;  et~al.
\newblock {Identification of Driver Genes for Critical Forms of COVID-19 in a
  Deeply Phenotyped Young Patient Cohort}.
\newblock {\em Science Translational Medicine} {\bf 2021}, {\em 14},~1--20.

\bibitem[Requejo-Castro et~al.(2020)Requejo-Castro, Gin{\'e}-Garriga, and
  P{\'e}rez-Foguet]{sdgs}
Requejo-Castro, D.; Gin{\'e}-Garriga, R.; P{\'e}rez-Foguet, A.
\newblock {Data-driven Bayesian Network Modelling to Explore the Relationships
  Between SDG 6 and the 2030 Agenda}.
\newblock {\em Science of The Total Environment} {\bf 2020}, {\em 710},~136014.

\bibitem[Zilko et~al.(2016)Zilko, Kurowicka, and Goverde]{railways}
Zilko, A.A.; Kurowicka, D.; Goverde, R.M.P.
\newblock {Modeling Railway Disruption Lengths with Copula Bayesian Networks}.
\newblock {\em Transportation Research Part C: Emerging Technologies} {\bf
  2016}, {\em 68},~350--368.

\bibitem[Gao et~al.(2020)Gao, Wang, Helu, and Teti]{cirp}
Gao, R.X.; Wang, L.; Helu, M.; Teti, R.
\newblock {Big Data Analytics for Smart Factories of the Future}.
\newblock {\em CIRP Annals} {\bf 2020}, {\em 69},~668--692.

\bibitem[Blei et~al.(2017)Blei, Kucukelbir, and McAuliffe]{blei}
Blei, D.M.; Kucukelbir, A.; McAuliffe, J.D.
\newblock {Variational Inference: A Review for Statisticians}.
\newblock {\em Journal of American Statistical Association} {\bf 2017}, {\em
  112},~859--877.

\bibitem[Dempster et~al.(1977)Dempster, Laird, and Rubin]{em}
Dempster, A.P.; Laird, N.M.; Rubin, D.B.
\newblock {Maximum Likelihood From Incomplete Data via the EM Algorithm}.
\newblock {\em Journal of the Royal Statistical Society (Series B)} {\bf 1977},
  {\em 39},~1--22.

\bibitem[Minka(2001)]{ep}
Minka, T.P.
\newblock {Expectation Propagation for Approximate Bayesian Inference}.
\newblock In Proceedings of the {17th Conference on Uncertainty in Artificial
  Intelligence (UAI)},  2001, pp. 362--369.

\bibitem[{van der Maaten} and Hinton(2008)]{tsne}
{van der Maaten}, L.; Hinton, G.
\newblock {Visualizing Data Using t-SNE}.
\newblock {\em Journal of Machine Learning Research} {\bf 2008}, {\em
  9},~2579--3605.

\bibitem[Becht et~al.(2019)Becht, McInnes, Healy, Dutertre, Kwok, Ng, Ginhoux,
  and Newell]{umap}
Becht, E.; McInnes, L.; Healy, J.; Dutertre, C.A.; Kwok, I.W.H.; Ng, L.G.;
  Ginhoux, F.; Newell, E.W.
\newblock {Dimensionality Reduction for Visualizing Single-Cell Data Using
  UMAP}.
\newblock {\em Nature Biotechnology} {\bf 2019}, {\em 37},~38--44.

\bibitem[Murphy(2022)]{pml1}
Murphy, K.P.
\newblock {\em {Probabilistic Machine Learning: An introduction}}; MIT Press,
  2022.

\bibitem[Murphy(2023)]{pml2}
Murphy, K.P.
\newblock {\em {Probabilistic Machine Learning: Advanced Topics}}; MIT Press,
  2023.

\bibitem[Moral et~al.(2021)Moral, Cano, and {G{\'o}mez-Olmedo}]{moral}
Moral, S.; Cano, A.; {G{\'o}mez-Olmedo}, M.
\newblock {Computation of Kullback-Leibler Divergence in Bayesian Networks}.
\newblock {\em Entropy} {\bf 2021}, {\em 23},~1122.

\bibitem[Hershey and Olsen(2007)]{gausmix}
Hershey, J.R.; Olsen, P.A.
\newblock {Approximating the Kullback Leibler Divergence Between Gaussian
  Mixture Models}.
\newblock In Proceedings of the {32nd IEEE International Conference on
  Acoustics, Speech and Signal Processing (ICASSP)},  2007, Vol.~IV, pp.
  317--320.

\bibitem[Beskos et~al.(2014)Beskos, Crisan, and Jasra]{beskos}
Beskos, A.; Crisan, D.; Jasra, A.
\newblock {On the Stability of Sequential Monte Carlo Methods in High
  Dimensions}.
\newblock {\em The Annals of Applied Probability} {\bf 2014}, {\em
  24},~1396--1445.

\bibitem[Scutari(2010)]{jss09}
Scutari, M.
\newblock {Learning Bayesian Networks with the bnlearn R Package}.
\newblock {\em Journal of Statistical Software} {\bf 2010}, {\em 35},~1--22.

\bibitem[Heckerman et~al.(1995)Heckerman, Geiger, and Chickering]{heckerman}
Heckerman, D.; Geiger, D.; Chickering, D.M.
\newblock {Learning Bayesian Networks: The Combination of Knowledge and
  Statistical Data}.
\newblock {\em Machine Learning} {\bf 1995}, {\em 20},~197--243.
\newblock Available as Technical Report MSR-TR-94-09.

\bibitem[Chickering and Heckerman(1994)]{nphard}
Chickering, D.M.; Heckerman, D.
\newblock {Learning Bayesian Networks is NP-Hard}.
\newblock Technical Report MSR-TR-94-17, Microsoft Corporation,  1994.

\bibitem[Chickering(1996)]{npcomp}
Chickering, D.M.
\newblock {Learning Bayesian Networks is NP-Complete}. In {\em Learning from
  Data: Artificial Intelligence and Statistics V}; Fisher, D.; Lenz, H., Eds.;
  Springer-Verlag,  1996; pp. 121--130.

\bibitem[Chickering et~al.(2004)Chickering, Heckerman, and Meek]{nplarge}
Chickering, D.M.; Heckerman, D.; Meek, C.
\newblock {Large-sample Learning of Bayesian Networks is NP-hard}.
\newblock {\em Journal of Machine Learning Research} {\bf 2004}, {\em
  5},~1287--1330.

\bibitem[Scutari et~al.(2019)Scutari, Vitolo, and Tucker]{stco17}
Scutari, M.; Vitolo, C.; Tucker, A.
\newblock {Learning Bayesian Networks from Big Data with Greedy Search:
  Computational Complexity and Efficient Implementation}.
\newblock {\em Statistics and Computing} {\bf 2019}, {\em 25},~1095--1108.

\bibitem[Cussens(2012)]{cutting}
Cussens, J.
\newblock {Bayesian Network Learning with Cutting Planes}.
\newblock In Proceedings of the {27th Conference on Uncertainty in Artificial
  Intelligence (UAI)},  2012, pp. 153--160.

\bibitem[Suzuki(2017)]{suzuki17}
Suzuki, J.
\newblock {An Efficient Bayesian Network Structure Learning Strategy}.
\newblock {\em New Generation Computing} {\bf 2017}, {\em 35},~105--124.

\bibitem[Scanagatta et~al.(2015)Scanagatta, de~Campos, Corani, and
  Zaffalon]{scanagatta}
Scanagatta, M.; de~Campos, C.P.; Corani, G.; Zaffalon, M.
\newblock {Learning Bayesian Networks with Thousands of Variables}.
\newblock {\em Advances in Neural Information Processing Systems (NIPS)} {\bf
  2015}, {\em 28},~1864--1872.

\bibitem[Hausser and Strimmer(2009)]{shrinkent}
Hausser, J.; Strimmer, K.
\newblock {Entropy Inference and the James-Stein Estimator, with Application to
  Nonlinear Gene Association Networks}.
\newblock {\em Journal of Machine Learning Resesearch} {\bf 2009}, {\em
  10},~1469--1484.

\bibitem[Agresti(2012)]{agresti}
Agresti, A.
\newblock {\em Categorical Data Analysis}, 3rd ed.; Wiley,  2012.

\bibitem[Geiger and Heckerman(1994)]{heckerman3}
Geiger, D.; Heckerman, D.
\newblock {Learning Gaussian Networks}.
\newblock In Proceedings of the {10th Conference on Uncertainty in Artificial
  Intelligence (UAI)},  1994, pp. 235--243.

\bibitem[Pourahmadi(2011)]{pourahmadi}
Pourahmadi, M.
\newblock {Covariance Estimation: The GLM and Regularization Perspectives}.
\newblock {\em Statistical Science} {\bf 2011}, {\em 26},~369--387.

\bibitem[Lauritzen and Wermuth(1989)]{lauritzen}
Lauritzen, S.L.; Wermuth, N.
\newblock {Graphical Models for Associations Between Variables, Some of which
  are Qualitative and Some Quantitative}.
\newblock {\em The Annals of Statistics} {\bf 1989}, {\em 17},~31--57.

\bibitem[Scutari et~al.(2022)Scutari, Marquis, and Azzimonti]{pgm22}
Scutari, M.; Marquis, C.; Azzimonti, L.
\newblock {Using Mixed-Effect Models to Learn Bayesian Networks from Related
  Data Sets}.
\newblock {\em Proceedings of Machine Learning Research (PGM)} {\bf 2022}, {\em
  186},~73--84.

\bibitem[Lauritzen and Spiegelhalter(1988)]{asia}
Lauritzen, S.L.; Spiegelhalter, D.J.
\newblock {Local Computation with Probabilities on Graphical Structures and
  their Application to Expert Systems (with discussion)}.
\newblock {\em Journal of the Royal Statistical Society: Series B (Statistical
  Methodology)} {\bf 1988}, {\em 50},~157--224.

\bibitem[Lauritzen and Jensen(2001)]{lauritzen2}
Lauritzen, S.L.; Jensen, F.
\newblock {Stable Local Computation with Conditional Gaussian Distributions}.
\newblock {\em Statistics and Computing} {\bf 2001}, {\em 11},~191--203.

\bibitem[Cowell(2005)]{cowell}
Cowell, R.G.
\newblock {Local Propagation in Conditional Gaussian Bayesian Networks}.
\newblock {\em Journal of Machine Learning Research} {\bf 2005}, {\em
  6},~1517--1550.

\bibitem[Namasivayam et~al.(2006)Namasivayam, Pathak, and Prasanna]{jtree-big0}
Namasivayam, V.K.; Pathak, A.; Prasanna, V.K.
\newblock {Scalable Parallel Implementation of Bayesian Network to Junction
  Tree Conversion for Exact Inference}.
\newblock In Proceedings of the {18th International Symposium on Computer
  Architecture and High Performance Computing},  2006, pp. 167--176.

\bibitem[Pennock(2023)]{pennock}
Pennock, D.M.
\newblock {Logarithmic Time Parallel Bayesian Inference}.
\newblock In Proceedings of the {14th Conference on Uncertainty in Artificial
  Intelligence (UAI)},  2023, pp. 431--438.

\bibitem[Namasivayam and Prasanna(2006)]{bp-big0}
Namasivayam, V.K.; Prasanna, V.K.
\newblock {Scalable Parallel Implementation of Exact Inference in Bayesian
  Networks}.
\newblock In Proceedings of the {12th International Conference on Parallel and
  Distributed Systems (ICPADS)},  2006, pp. 1--8.

\bibitem[Malioutov et~al.(2006)Malioutov, Johnson, and Willsky]{malioutov}
Malioutov, D.M.; Johnson, J.K.; Willsky, A.S.
\newblock {Walk-Sums and Belief Propagation in Gaussian Graphical Models}.
\newblock {\em Journal of Machine Learning Research} {\bf 2006}, {\em
  7},~2031--2064.

\bibitem[Cheng and Druzdzel(2000)]{aisbn}
Cheng, J.; Druzdzel, M.J.
\newblock {AIS-BN: An Adaptive Importance Sampling Algorithm for Evidential
  Reasoning in Large Bayesian Networks}.
\newblock {\em Journal of Artificial Intelligence Research} {\bf 2000}, {\em
  13},~155--188.

\bibitem[Yuan and Druzdzel(2003)]{episbn}
Yuan, C.; Druzdzel, M.J.
\newblock {An Importance Sampling Algorithm Based on Evidence Pre-Propagation}.
\newblock In Proceedings of the {19th Conference on Uncertainty in Artificial
  Intelligence (UAI)},  2003, pp. 624--631.

\bibitem[Cover and Thomas(2006)]{itheory}
Cover, T.M.; Thomas, J.A.
\newblock {\em {Elements of Information Theory}}, 2nd ed.; Wiley,  2006.

\bibitem[Csisz{\'a}r and Shields(2004)]{csiszar}
Csisz{\'a}r, I.; Shields, P.
\newblock {\em {Information Theory and Statistics: A Tutorial}}; Now Publishers
  Inc.,  2004.

\bibitem[G{\'o}mez-Villegas et~al.(2008)G{\'o}mez-Villegas, Main, and
  Susi]{gbnkl1}
G{\'o}mez-Villegas, M.A.; Main, P.; Susi, R.
\newblock {Sensitivity of Gaussian Bayesian Networks to Inaccuracies in Their
  Parameters}.
\newblock In Proceedings of the {4th European Workshop on Probabilistic
  Graphical Models (PGM)},  2008, pp. 265--272.

\bibitem[G{\'o}mez-Villegas et~al.(2013)G{\'o}mez-Villegas, Main, and
  Susi]{gbnkl2}
G{\'o}mez-Villegas, M.A.; Main, P.; Susi, R.
\newblock {The Effect of Block Parameter Perturbations in Gaussian Bayesian
  Networks: Sensitivity and Robustness}.
\newblock {\em Information Sciences} {\bf 2013}, {\em 222},~439--458.

\bibitem[G{\"o}rgen and Leonelli(2020)]{gbnkl3}
G{\"o}rgen, C.; Leonelli, M.
\newblock {Model-Preserving Sensitivity Analysis for Families of Gaussian
  Distributions}.
\newblock {\em Journal of Machine Learning Research} {\bf 2020}, {\em
  21},~1--32.

\bibitem[Seber(2008)]{seber}
Seber, G.A.F.
\newblock {\em {A Matrix Handbook for Stasticians}}; Wiley,  2008.

\bibitem[Stewart(1998)]{stewart}
Stewart, G.W.
\newblock {\em {Matrix Algorithms, Volume I: Basic Decompositions}}; SIAM,
  1998.

\bibitem[Cavanaugh(2004)]{cavanaugh}
Cavanaugh, J.E.
\newblock {Criteria for Linear Model Selection Based on Kullback's Symmetric
  Divergence}.
\newblock {\em Australian \& New Zealand Journal of Statistics} {\bf 2004},
  {\em 46},~197--323.

\end{thebibliography}

\PublishersNote{}
\end{adjustwidth}
\end{document}